\documentclass[10pt,twocolumn,letterpaper]{article}

\usepackage[pagenumbers]{cvpr} 

\usepackage{graphicx}
\usepackage{amsmath}
\usepackage{amssymb}
\usepackage{booktabs}
\usepackage{multirow}
\usepackage[table,xcdraw]{xcolor}

\usepackage[pagebackref,breaklinks,colorlinks]{hyperref}

\usepackage[capitalize]{cleveref}
\crefname{section}{Sec.}{Secs.}
\Crefname{section}{Section}{Sections}
\Crefname{table}{Table}{Tables}
\crefname{table}{Tab.}{Tabs.}

\begin{document}

\title{TextFormer: A Query-based End-to-End Text Spotter with Mixed Supervision}

\author{Yukun Zhai$^1$, Xiaoqiang Zhang$^2$, Xiameng Qin$^2$, Sanyuan Zhao$^1$, Xingping Dong$^1$, Jianbing Shen$^1$ \\
$^1${School of Computer Science}, {Beijing Institute of Technology} \\
$^2${Department of Computer Vision Technology}, {Baidu Inc.} \\
{\tt\small \{zhaiyukun, zhaosanyuan, shenjianbing\}@bit.edu.cn} \\
{\tt\small \{xingping.dong\}@gmail.com, \{zhangxiaoqiang01, qinxiameng\}@baidu.com}
}

\maketitle

\begin{abstract}
End-to-end text spotting is a vital computer vision task that aims to integrate scene text detection and recognition into a unified framework. Typical methods heavily rely on Region-of-Interest (RoI) operations to extract local features and complex post-processing steps to produce final predictions. To address these limitations, we propose TextFormer, a query-based end-to-end text spotter with Transformer architecture. Specifically, using query embedding per text instance, TextFormer builds upon an image encoder and a text decoder to learn a joint semantic understanding for multi-task modeling. It allows for mutual training and optimization of classification, segmentation, and recognition branches, resulting in deeper feature sharing without sacrificing flexibility or simplicity. Additionally, we design an Adaptive Global aGgregation (AGG) module to transfer global features into sequential features for reading arbitrarily-shaped texts, which overcomes the sub-optimization problem of RoI operations. Furthermore, potential corpus information is utilized from weak annotations to full labels through mixed supervision, further improving text detection and end-to-end text spotting results. Extensive experiments on various bilingual (i.e., English and Chinese) benchmarks demonstrate the superiority of our method. Especially on TDA-ReCTS dataset, TextFormer surpasses the state-of-the-art method in terms of 1-NED by 13.2\%.
\end{abstract}

\section{Introduction}
\label{sec:intro}

Scene text spotting, which aims to locate and read text from natural images, has received widespread academic and industrial attention due to its numerous real-world application values, including image retrieval~\cite{datta2008image}, fake news detection~\cite{reddy2020text}, augmented reality~\cite{wu2019editing}, blind navigation~\cite{rong2016guided}, visual question answering~\cite{antol2015vqa}, document understanding~\cite{li2021structext,yu2023structextv2,zhai2023fast}, etc.
Traditionally, a scene text spotter consists of a text detector and a recognizer. The detector identifies the position of text, represented as polygon coordinates or a shape mask, and the recognizer transcribes the corresponding text.
Unlike isolation-based training for text detection and recognition tasks~\cite{jaderberg2016reading,gomez2017textproposals,neumann2015real}, end-to-end methods~\cite{lyu2018mask,liu2020abcnet,wang2021pan++, qin2019towards, qiao2021mango, qiao2020text, wang2020all, xing2019convolutional, wang2021pgnet} that train two tasks within a unified framework have received increasing interest due to their benefits of a unified framework and global optimization. 
However, there is still room for improvement due to the inherent challenges of end-to-end text spotting, such as arbitrarily-shaped texts with large variations in font, style, and size and ambiguous texts with juxtaposed text lines~\cite{wang2020ae}.

\begin{figure*}[t]
\centering
\includegraphics[width=0.94\linewidth]{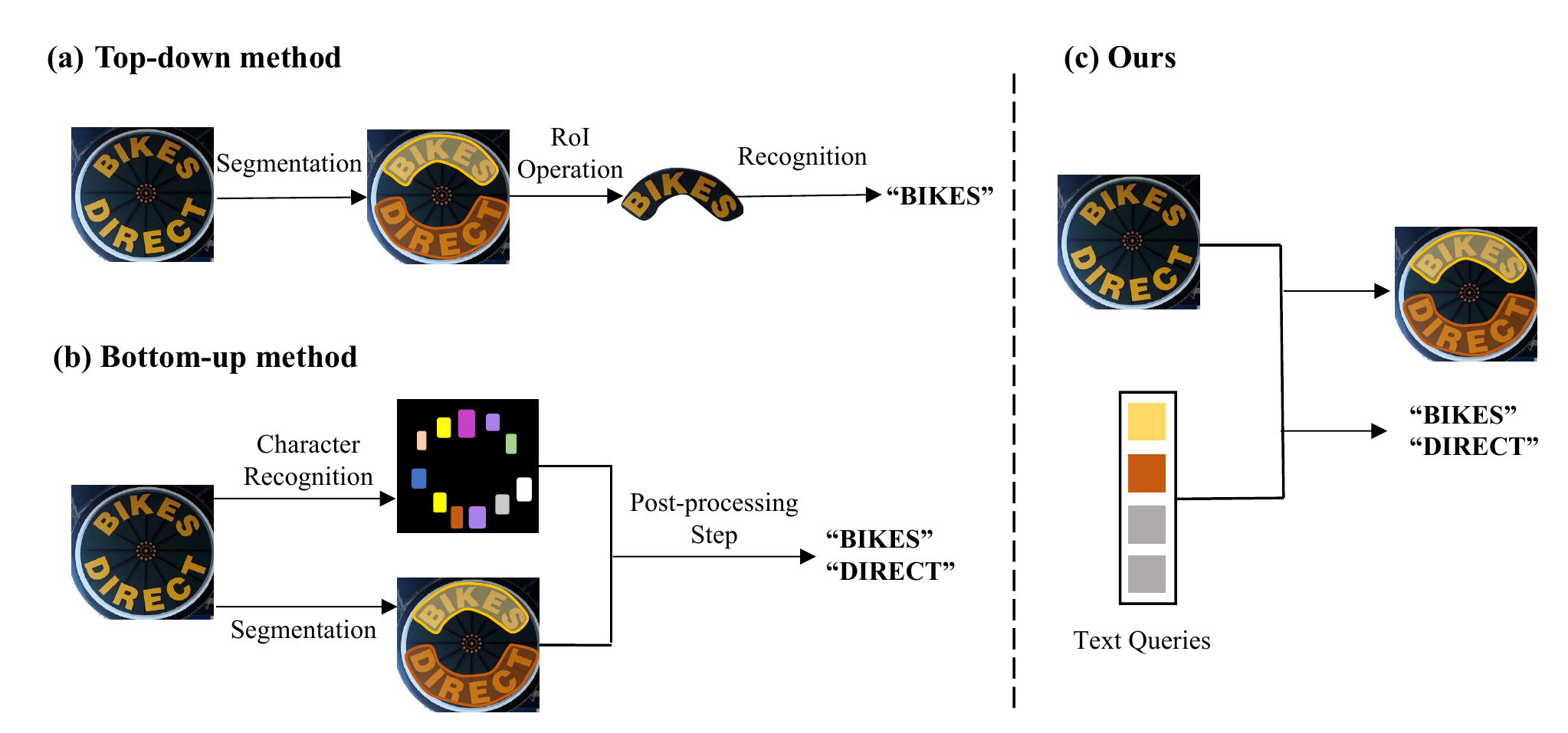}
\caption{Comparison of previous end-to-end text spotting methods with our query-based pipeline. (\textbf{a}) shows the traditional top-down (two-stage) framework, which uses RoI operation to extract features and integrate the detection and recognition branches. (\textbf{b}) depicts the bottom-up framework (one-stage), which predicts all possible characters and requires intricate post-processing step to match all characters into the correct words. (\textbf{c}) is our query-based framework that directly outputs both segmentation and recognition results without RoI operation and post-processing step.}
\label{fig1}
\end{figure*}

Early end-to-end methods~\cite{liu2020abcnet, wang2021pan++, qin2019towards, qiao2020text, wang2020all, liu2018fots} are designed using a top-down approach, following a two-stage paradigm that first detects text areas and then recognizes corresponding text using Region-of-Interest (RoI) operation, as depicted in Fig.~\ref{fig1} (a). Despite their promising performance, these methods only share shallow backbone features and adopt local RoI features for different tasks, which can lead to two significant problems: (1) The recognition performance is heavily reliant on the accuracy of detection results; (2) The scope of recognition learning is limited to enhancing detection features. As illustrated in Fig.~\ref{fig1} (b), recent bottom-up methods~\cite{xing2019convolutional,qiao2021mango,wang2021pgnet} strive to break free from the top-down paradigm by involving character detection and classification, thus reducing the strong dependence between text detection and recognition. Yet these methods are plagued by inaccurate character detection results and require time-costly character-level annotation.

People naturally focus on the general area while reading the text and then adjust their focus to find the correct content~\cite{ricoeur1971model}. Recognizing this pattern and the limitations of the above frameworks, we propose there is no strong coupling between text detection and recognition tasks, and they can be mutually promoted by sharing deeper features through a multi-task model design. And inspired by recent query-based methods ~\cite{carion2020end, zhu2020deformable, cheng2022masked} in object detection and instance segmentation, we introduce TextFormer, a query-based end-to-end text spotter based on Transformer architecture. It comprises an image encoder, a text decoder, and a multi-task module containing classification, segmentation, and recognition branches. For each text query, TextFormer predicts its category, segmentation mask, and corresponding text transcription in parallel. The multi-task module can be collaboratively trained and optimized to share complementary information by learning shared semantic features from the text decoder.
In addition, an Adaptive Global feature aGgregation (AGG) module is employed to extract features of different orientations for text recognition, enabling our network to read arbitrarily-shaped text. 

\begin{figure}[htbp]
\centering
\includegraphics[width=0.94\linewidth]{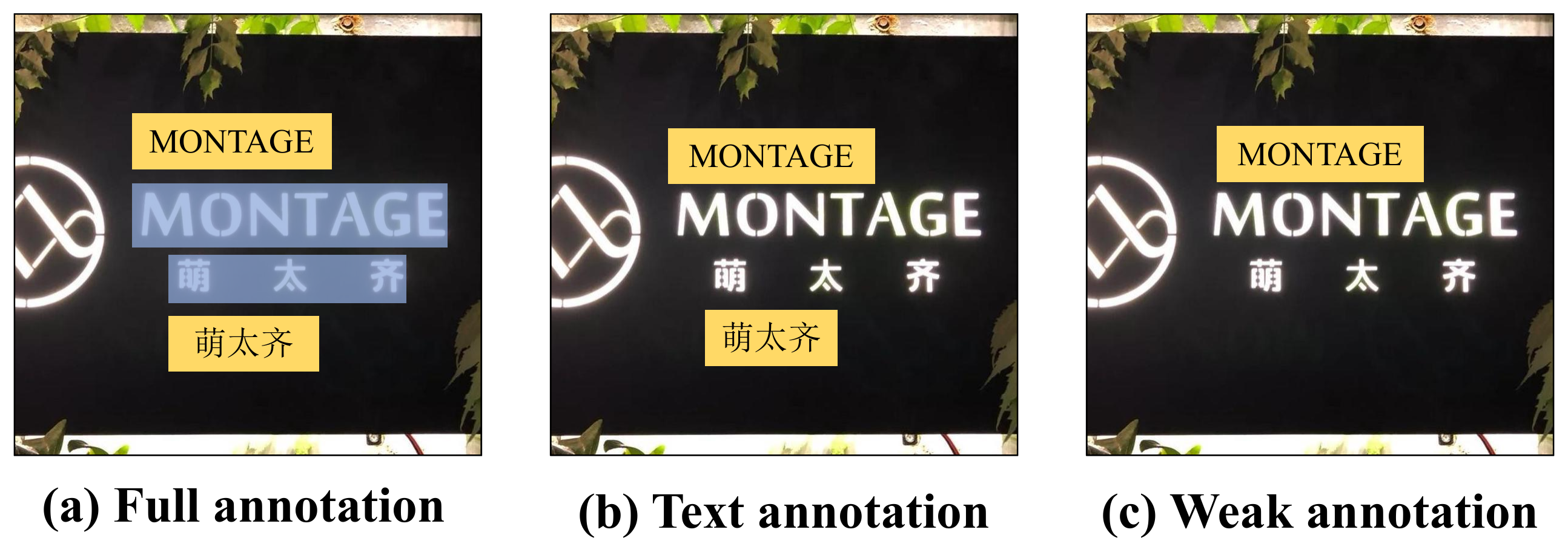}
\caption{Examples of images with different annotations. (\textbf{a}) is the full annotation that includes text regions with text transcription labels. (\textbf{b}) shows the text annotation that contains text transcription annotations without location information. (\textbf{c}) indicates the weak annotation that only the transcription of text-of-interest is provided. The regions commonly referred to as ``Text-of-interest" typically contain the names of store fonts or descriptions of landmarks, supplying significant information for localization and navigation purposes~\cite{sun2019icdar}.}
\label{fig2}
\end{figure}

However, achieving adequate joint training is not straightforward: (1) Full annotations (Fig.~\ref{fig2} (a)) to support end-to-end text spotting are overly expensive; (2) The recognition branch requires far more training data than the detection branch, especially for reading Chinese text in the wild. This is because the number of Chinese characters is much larger than that of Latin-based languages. Previous methods solve this problem by replacing time-costly bounding box or polygon annotations with single-point annotations~\cite{peng2022spts} or even retaining only text annotations~\cite{tang2022you} (Fig.~\ref{fig2} (b)), while end-to-end text spotting performance is largely behind compared to the same text spotter trained with full annotations. 

To address this issue, we present mixed supervision for training an end-to-end text spotter. It utilizes a mixture of weak annotations (Fig.~\ref{fig2} (c)) and full annotations, which can meet the needs of data quantity for text recognition and further boost the performance of end-to-end text spotting. And in order to allow this training setting, we simply modify the original composition of Hungarian matching loss~\cite{carion2020end} without changing the network structure. With the generic design of our framework and mixed-supervision training strategy, our TextFormer achieves the state-of-the-art performance on common benchmarks.

In summary, our main contributions can be summarized as follows: 
\begin{itemize}
\item   We design a novel, fully end-to-end text spotter with a multi-task model design that uses text queries to bridge the classification, segmentation, and recognition branches together. It also allows a global feature extractor named AGG to extract features from different orientations for reading arbitrarily-shaped texts.
\item   We train our network with mixed supervision to improve the effect of co-optimization for text detection and recognition, which utilizes a mixture of weak annotation and full labels. To our knowledge, we are the first to adopt mixed supervision to tackle the end-to-end text spotting task.
\item   Extensive experiments on public benchmarks demonstrate the state-of-the-art performance of our method on both text detection and end-to-end text spotting tasks. In particular, we evaluated TextFormer on an ambiguous text spotting dataset named TDA-ReCTS, where it outperforms its counterparts by a large margin of 13.2\% in terms of 1-NED.
\end{itemize}

\section{Related Work}

\subsection{End-to-end Text Spotting}
We compare two paradigms for typical end-to-end text spotting frameworks, top-down and bottom-up methods. The top-down approach employs RoI operations to link the text detector and recognizer. In contrast, the bottom-up approach utilizes character as the basic unit to perform text detection and recognition, eliminating the need for the RoI step. 
\subsubsection{Top-down Methods}
The initial approaches aim to read regular texts via a top-down framework. The first end-to-end trainable text spotter~\cite{li2017towards} was inspired by Faster R-CNN~\cite{ren2015faster} and used RoI Pooling to unify the detection and recognition parts into a unified network. An advanced version~\cite{busta2017deep} was designed to handle multi-oriented texts. To further integrate the detection and recognition stages,  FOTS~\cite{liu2018fots} introduced RoI-Rotate to extract the text proposal features corresponding to the detection results. Meanwhile, He et al.~\cite{he2018end} developed a similar framework that employed an attention-based sequence decoder as its recognition head. Despite the excellent performance of these methods on regular texts, these methods were ineffective at processing arbitrarily-shaped texts.

Based on Mask R-CNN~\cite{he2017mask}, Mask TextSpotter~\cite{lyu2018mask} added a character-level segmentation branch to read the text of arbitrary shapes. Qin et al.~\cite{qin2019towards} directly adopted Mask R-CNN and introduced RoI Masking to crop features with segmentation masks. 
And PAN++~\cite{wang2021pan++} proposed an optimized feature extractor named Masked RoI to better remove noise from the background.
TextDragon~\cite{feng2019textdragon} described the shape of text with a series of quadrangles and used RoISlide to sample features. Boundary~\cite{wang2020all} and Text Perceptron~\cite{qiao2020text} regraded the text boundaries as key points and applied the thin-plate-spline transformation~\cite{bookstein1989principal} to rectify irregular text instances. ABCNet~\cite{liu2020abcnet} and ABCNet v2~\cite{liu2021abcnetv2} fitted arbitrarily-shaped texts by parameterized Bezier curves and proposed BezierAlign as the feature extractor. 

Most of the methods mentioned above use RoI operations to connect the text detector and the recognizer in a linear process of detection followed by recognition. As a result, the recognition performance is depended on the accuracy of detection results and RoI operations, especially for ambiguous texts~\cite{wang2020ae}.

\subsubsection{Bottom-up Methods}
Recently, some works have attempted to overcome the limitations of RoI operations for top-down methods. CharNet~\cite{xing2019convolutional} directly segmented single character instance, which requires character-level annotations for training. And the final text instance can be generated by post-processing steps to group characters into words. MANGO~\cite{qiao2021mango} exploited the position-ware mask attention module to predict characters in each text instance, which utilized the sequence context information among individual characters. To avoid using character-level annotations, PGNet~\cite{wang2021pgnet} learned the pixel-level character classification map by point-gathering operation, while the text recognition module was still dependent on the detection part. The above approaches alleviate the reliance between detection and recognition breaches. However, character-level labels or complex post-processing steps are still required for training and inference.

\begin{figure*}[h!t]
\centering
\includegraphics[width=1\linewidth]{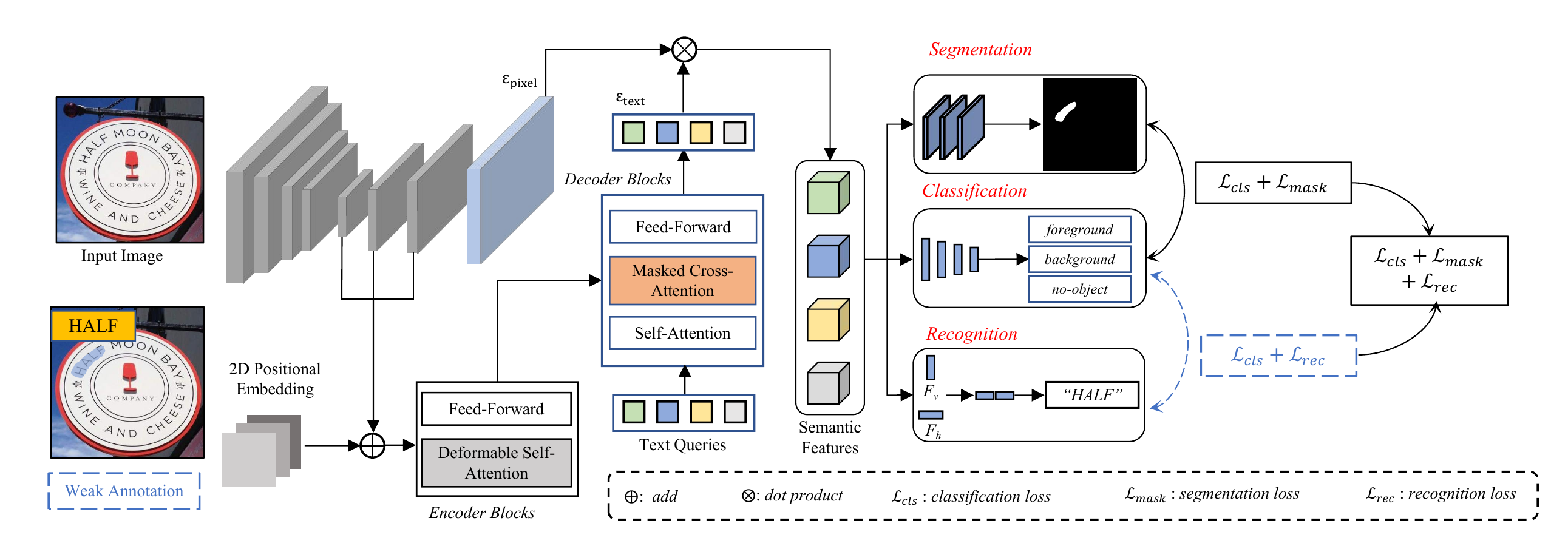}
\caption{The overall architecture of TextFormer. We propose a query-based end-to-end text spotter where each query represents a latent text instance. Our mixed-supervision training scheme allows our network to train with a mix of fully and weakly labeled data for better results. Note the composition of the loss function for training weakly supervised data is framed by the dashed line.}
\label{fig3}
\end{figure*}

\subsubsection{Query-based Methods}
DETR~\cite{carion2020end} has introduced a new paradigm in object detection and image segmentation, eliminating the need for RoI operators and complex post-processing procedures. Instead, object queries are used to aggregate features and then generate the detection and segmentation results for each object instance using the Transformer encoder-decoder architecture. Researchers have attempted to lead this paradigm to the scene text detection task with tailored designs. Raisi et al.~\cite{raisi2021transformer} was the first to introduce the Transformer framework for scene text detection, treating text instances as object queries to handle multi-oriented text. Jingqun et al.~\cite{tang2022few} then proposed a novel encoder-decoder architecture that performed text detection based on a few sampled point features represented by feature queries. 

Our TextFormer is a scene text spotter based on the query design.
Specifically, it leverages text queries to represent all possible text in the scene image and character queries to represent the characters within each text instance. 

\subsection{Mixed Supervision}
Previous works have focused on using varying levels of annotated data in computer vision tasks, such as object detection~\cite{bovzivc2021mixed}, segmentation~\cite{mlynarski2019deep}, and scene text detection~\cite{tian2017wetext}, etc. SPTS~\cite{peng2022spts} used single-point labels to indicate text locations, while TOSS~\cite{tang2022you} utilized a fully transcription-based supervised approach that trained the text spotter with text annotations. Despite using different kinds of annotated data, the model performance is largely behind the common text spotter. Instead, we use a mix of weak annotations with full annotations to leverage the potential of the query-based text spotter and further enhances the performance of our model, which is formulated as mixed supervision. To the best of our knowledge, we are the first to adopt mixed supervision to tackle the end-to-end text spotting task.

\section{Methodology}

Our TextFormer consists of four main components: (1) an image encoder-decoder to extract semantic features and share them with the multi-task branches, (2) a classification branch to classify text queries, (3) a segmentation branch to segment text areas, and (4) a recognition branch to read text. In this section, we first present the framework of TextFormer and then introduce our mixed supervision tailored for end-to-end text spotting. 

\subsection{Overall Architecture} 
The overall architecture of TextFormer is illustrated in Fig.~\ref{fig2}. The encoder blocks perform self-attention across multi-scale feature maps. Then, text queries are fed through the decoder blocks to generate semantic features for the parallel heads, which predict masks and corresponding characters to each text instance.
We then specify the implementations of each module in the sub-sections below.

\begin{figure*}[h!t]
\centering
\includegraphics[width=0.95\linewidth]{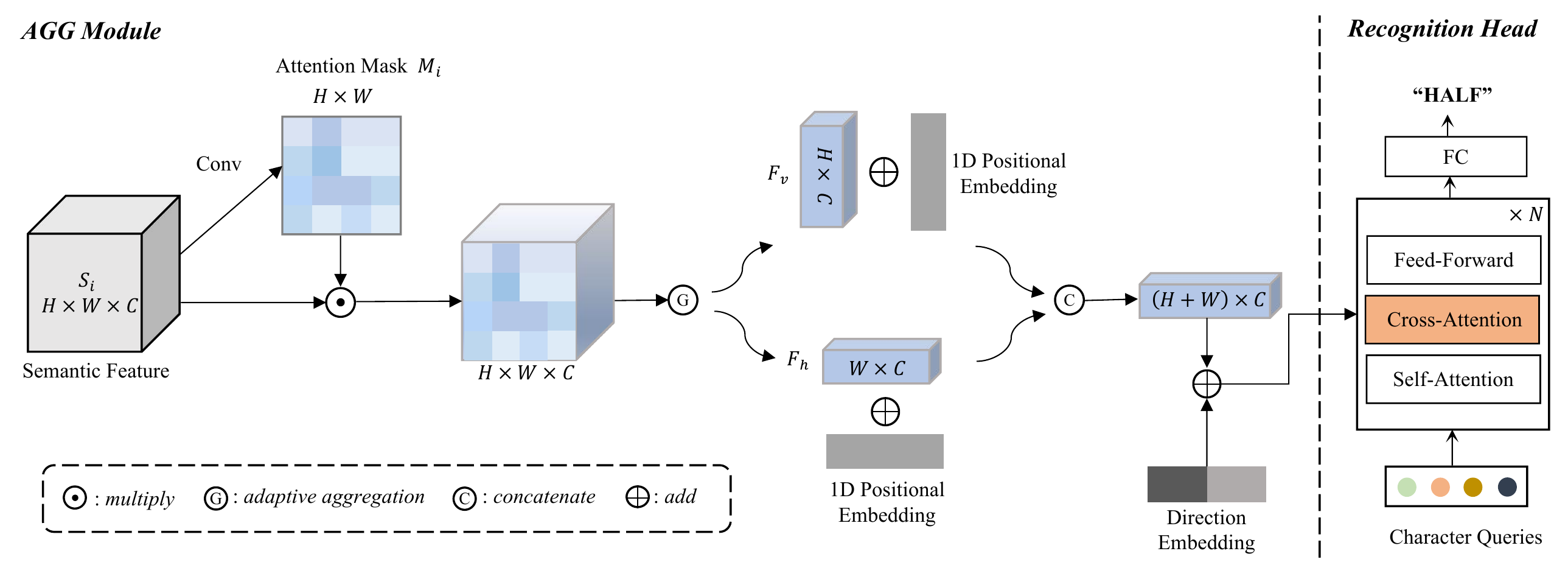}
\caption{The detailed structure of our recognition branch. It is composed of a feature extractor and an attention-based recognizer. The AGG module extracts the sequential features from the shared semantic features. The Transformer-based recognition head predicts characters for the corresponding character queries.}
\label{fig4}
\end{figure*}

\subsection{Encoder-Decoder Module}
Given a natural scene image $X\in\mathbb{R}^{H\times W\times 3}$, the backbone with Feature Pyramid Network (FPN) ~\cite{lin2017feature} extracts feature maps $P_5$,$P_4$,$P_3$,$P_2$. 
The resolution of multi-scale feature maps correspond to $\frac{1}{32},\frac{1}{16},\frac{1}{8},\frac{1}{4}$ comparing to original image size with 256 channels respectively. 
We then flatten and concatenate the first three features into feature sequence with size of $(L_5 + L_4 + L_3) \times 256$, where $L_i$ denotes the flattened length of $P_i$, which is $\frac{H}{2^i} \times \frac{W}{2^i}$. Added with 2D Positional Embedding~\cite{carion2020end}, the feature sequence is fed into Transformer encoder blocks to generate refined features. We equip each encoder layer with deformable attention~\cite{zhu2020deformable} that constraints cross-attention at a fixed number of reference points, which can largely reduce computational overheads and better capture small text instances. 

We take $N$ randomly initialized embeddings as text queries, standing for each possible text instance in the scene image. Next, using text queries and refined features as inputs, the Transformer decoder blocks 
output $N$ text embeddings $\varepsilon_{text}\in\mathbb{R}^{N\times256}$. To speed up the convergence of the training process, we replace the standard Transformer decoder with masked-attention decoder~\cite{cheng2022masked}, which limits cross attention to within the predicted region for each text query.

After getting text embeddings $\varepsilon_{text}$ from decoder blocks, we compute the shared semantic features $S$ for multi-task heads. Concretely, we first upsample the $P_2$ feature to generate pixel embedding $\varepsilon_{pixel}\in\mathbb{R}^{\frac{H}{4}\times\frac{W}{4}\times 256}$. Then we obtain each semantic feature $S_i\in\mathbb{R}^{\frac{H}{4}\times\frac{W}{4}\times 256}$ via a dot product between text embedding $\varepsilon_{text}$ and pixel embedding $\epsilon_{pixel}$. The shared semantic features contain both global and local information, which are devoted to mutually optimizing the network.

\subsection{Recognition Branch} 
The detailed structure of our recognition branch is shown in Fig.~\ref{fig4}. It is composed of a global feature extractor and an attention-based recognizer.


\subsubsection{AGG Module}
The AGG module serves as a global feature extractor, which is used to obtain sequential features from the shared semantic features. 
Our approach is inspired by the observed linearity of the line-by-line reading order, and as such, we aggregate the global feature map in both horizontal and vertical orientations. These aggregated features are then concatenated and used for scene text recognition.

Firstly, we employ a $1\times1$ convolution layer with a sigmoid function to compute the attention mask $M_i\in\mathbb{R}^{\frac{H}{4}\times\frac{W}{4}\times256}$ of each text query from the shared semantic feature $S_i$, which is described as:
\begin{align}
M_{i} = sigmoid(Conv(S_i)) \label{eq1}
\end{align}

Secondly, as shown in Fig.~\ref{fig4}, we multiply the shared semantic feature $S_i$ by the attention mask $M_i$ to pay more attention to the text instance and decrease the effect of background noise. Then, we adaptive aggregating the obtained global feature map under the guidance of the attention mask $M_i$ in the corresponding column to obtain the horizontal feature $F_h$. The vertical feature $F_v$ can be extracted in the same way. Thus, adaptive global feature aggregation could be formulated in two directions as:
\begin{equation}
\begin{aligned}
F_h = AGG_h(S_i \odot W_i) \\
F_v = AGG_v(S_i \odot W_i)
\end{aligned}
\end{equation}
where $\odot$ denotes element-wise multiplication. $ AGG_{(\cdot)}(\cdot) $ represents an adaptive summation operation in the corresponding direction,  as implemented in our model. Specifically, we compute the sum of the feature map along a certain direction and divide it by the sum of the mask attention map in the same direction. As such, these two directional vectors contain information pertaining to the character sequence in the different orientations.

Finally, we obtain the sequence vector fed into the recognition head as:
\begin{align}
x_{seq} = concat\left[F_h \oplus e_h, F_v^T \oplus e_v \right ] \oplus e_d
\end{align}
In addition to the 1D sine positional embedding added to $F_h$ and $F_v$, we add a direction embedding, denoted as $e_d$, to the concatenated feature sequence, which allows the following recognition head to identify the direction in each feature sequence lies. The direction embedding is randomly initialized and can be jointly trained in our network.

The proposed AGG module is completely differentiable, enabling the recognition loss to be propagated to the segmentation branch, thereby allowing mutual training and optimization of both branches in an end-to-end manner. Moreover, the AGG module extracts features directly from the global feature maps without the need for any spatial rectification operation.

\subsubsection{Recognition Head}
Since the input sequence vector $x_{seq}$ is already rich in information after being encoded by the AGG module, we simply use the standard Transformer-based decoder as our recognition head.
The initial character queries are randomly initialized, similar to the above-mentioned text queries. With the output sequence of the Transformer decoder, we can transform the recognition task into a parallel character classification problem. Concretely, we employ a fully-connected (FC) layer to get the final predicted character sequence.
\begin{align}
x_{rec} = FC(Decoder(x_{seq}))
\end{align}
where $x_{rec} \in \mathbb{R}^{K \times C}$. $K$ is the number of character queries and $C$ is the number of character classes. 
During training, the ground truth text is padded with $\left[\text{PAD}\right]$ symbols until reaches the pre-defined length $K$.

\subsection{Classification Branch}
The classification branch aims to classify the text queries from semantic features. We use global average pooling and three 
consecutive FC layers with a softmax activation function to predict class probabilities which include text, background, and no-text classes. 

\subsection{Segmentation Branch}
The segmentation head of our model is designed using a stack of $3\times 3$ convolution layers, followed by a $1\times 1$ convolution layer that incorporates a sigmoid function to obtain the final text instance mask. To avoid overlap, each pixel is assigned to the instance with the highest mask probability by merging all of these segmentation masks. Ultimately, the polygons of each instance can be generated by applying connected component analysis to the corresponding text mask.

\subsection{Mixed Supervision}\label{mixed}
Our TextFormer benefits from the parallel design, as illustrated in Fig.~\ref{fig3}, which allows for simultaneous multi-task learning based on the shared semantic features. However, while detecting text position is relatively straightforward, training text recognition requires a significantly larger amount of data and training rounds~\cite{qin2019towards}. To improve the co-optimization between text detection and recognition, it is important to make the best use of the different annotated labels. To address this problem, we propose mixed supervision to reduce the inconsistency between text detection and recognition. During this training paradigm, hybrid data with various annotations are fed into our network, as shown in Fig.~\ref{fig2}. Full annotations involve both text instance masks and corresponding transcriptions. Text annotations provide instance-level transcriptions, while weak annotations provide a single transcription of the text of interest~\cite{sun2019icdar}.

\subsubsection{Matching Cost}
Assuming that the ground-truth of text instance is represented as a triplet set $(C_q, M_q, T_q)$ which indicates query class, query mask, and text transcription. For convenience, we describe the text transcription as $T_q = \left \{ t^{k}_q \right \}^K_{k=1}$, where $K$ is the length of character sequence. The predicted result $\hat{y}$ and the label $y$ are formulated as:
\begin{equation}
\begin{aligned}
& \hat{y}= \left \{ \hat{y}_j = (\hat{C}_j, \hat{M}_j, \hat{T}_j) \right \}_{j=1}^{N} \\
& y = \left \{ \left \{ y_{q} = (C_q, M_q, T_q) \right \}_{q=1}^{N_{gt}}, \left \{\phi, \cdots, \phi \right \}^{N - N_{gt}} \right \}
\end{aligned}
\end{equation}
where $N$ is a pre-defined number of text predictions in our method, which equals to the number of text queries. Moreover, we append several $\phi$ tags (meaning ``no text") to pad the the number of text instance labels in ground truth to $N$.

In order to train our network with different supervisions, we employ a bipartite matching strategy~\cite{carion2020end} to compute the one-to-one matching cost. The matching cost contains three components: classification cost, mask cost, and recognition cost. Suppose that $\{ y_q, \hat{y}_j \} $ is the candidate of matched pair, the matching cost can be described as:
\begin{equation}
\begin{aligned}
\mathcal{L}_{match}(y_q,\hat{y}_j) = \{
\mathcal{C}_{mask}(M_q, \hat{M}_{j}) - \mathcal{C}_{rec}(T_q, \hat{T}_{j}) \\
- \mathcal{C}_{cls}(C_q)\}_{\{C_q     \neq \phi\}}
\end{aligned}
\end{equation}
where $\mathcal{C}_{cls}$ is the classification cost that equals the predicted probability of query class. $\mathcal{C}_{mask}$ indicates the mask cost which is the similarity of the predicted and ground truth masks. The recognition cost $\mathcal{C}_{rec}$ calculates the average probability of predicted text sequence with ground truth, which is represented as:
\begin{align}
\mathcal{C}_{rec}(T_q, \hat{T}_{j}) = \sum_{k=1}^{K} \hat{t}_{j}^k(t_{q}^k) 
\end{align}

At last, we obtain the best one-to-one matching pairs by minimizing the total matching cost $\mathcal{L}_{match}$, which is expressed as:
\begin{align}
\hat{\sigma }= \underset{ \sigma \in \mathfrak{S}_{N}}{argmin}\sum_{q}^{N}\mathcal{L}_{match}(y_q,\hat{y}_j)  
\end{align}
where $ \mathfrak{S}_{N} $ are the permutations sets for matching.

\begin{table*}[h!t]
\center
\caption{End-to-end text spotting results on ICDAR 2015. ``S", ``W", ``G" mean text recognition with Strong (S), Weak (W), and Generic (G) lexicon, respectively. The prefix ``S-" and ``L-" indicate rescaling the input image by the shorter or longer side. MANGO*~\cite{qiao2021mango} is evaluated with IOU 0.1. \label{tab1}}
\begin{tabular}{llllllll}
\toprule
\multirow{2}{*}{Method}	& \multirow{2}{*}{Input Size} & \multicolumn{3}{c}{Detection} & \multicolumn{3}{c}{End-to-end}  \\
\cmidrule(lr){3-5} \cmidrule(lr){6-8} 
 & & Precision & Recall & F-measure & \multicolumn{1}{c}{S} & \multicolumn{1}{c}{W} & \multicolumn{1}{c}{G} \\
\midrule
FOTS~\cite{liu2018fots} & L-2240 & \multicolumn{1}{c}{91.0} & \multicolumn{1}{c}{85.2} & \multicolumn{1}{c}{88.0} & \multicolumn{1}{c}{81.1} & \multicolumn{1}{c}{75.9} & \multicolumn{1}{c}{60.8}  \\
TextNet~\cite{sun2019textnet} & - & \multicolumn{1}{c}{89.4} & \multicolumn{1}{c}{85.4} & \multicolumn{1}{c}{87.4} & \multicolumn{1}{c}{78.7} & \multicolumn{1}{c}{74.9} & \multicolumn{1}{c}{60.5}  \\
He \textit{et al.}~\cite{he2018end} & - & \multicolumn{1}{c}{87.0} & \multicolumn{1}{c}{86.0} & \multicolumn{1}{c}{87.0} & \multicolumn{1}{c}{82.0} & \multicolumn{1}{c}{77.0} & \multicolumn{1}{c}{63.0}  \\
Mask TextSpotter~\cite{lyu2018mask} & S-1600 & \multicolumn{1}{c}{91.6} & \multicolumn{1}{c}{81.0} & \multicolumn{1}{c}{86.0} & \multicolumn{1}{c}{79.3} & \multicolumn{1}{c}{73.0} & \multicolumn{1}{c}{62.4}  \\
TextDragon ~\cite{feng2019textdragon} & - & \multicolumn{1}{c}{\underline{92.5}} & \multicolumn{1}{c}{83.8} & \multicolumn{1}{c}{87.9} & \multicolumn{1}{c}{82.5} & \multicolumn{1}{c}{78.3} & \multicolumn{1}{c}{65.1}  \\
Unconstrained ~\cite{qin2019towards} &S-900 &\multicolumn{1}{c}{89.4} &\multicolumn{1}{c}{85.8} &\multicolumn{1}{c}{87.5} &\multicolumn{1}{c}{83.4} &\multicolumn{1}{c}{79.9} &\multicolumn{1}{c}{68.0}  \\ 
Text Perceptron~\cite{qiao2020text} & L-2000 & \multicolumn{1}{c}{92.3} & \multicolumn{1}{c}{82.5} & \multicolumn{1}{c}{87.1} & \multicolumn{1}{c}{80.5} & \multicolumn{1}{c}{76.6} & \multicolumn{1}{c}{65.1}  \\
Mask TextSpotter v3~\cite{liao2020mask} &S-1440 &\multicolumn{1}{c}{-} &\multicolumn{1}{c}{-} & \multicolumn{1}{c}{-} &\multicolumn{1}{c}{83.3} &\multicolumn{1}{c}{78.1} &\multicolumn{1}{c}{74.2} \\ 
PAN++~\cite{wang2021pan++}  & S-896 & \multicolumn{1}{c}{91.4} & \multicolumn{1}{c}{83.9} & \multicolumn{1}{c}{87.5} & \multicolumn{1}{c}{82.7} & \multicolumn{1}{c}{78.2} & \multicolumn{1}{c}{69.2} \\ 
ABCNet v2~\cite{liu2021abcnetv2} & S-1000 & \multicolumn{1}{c}{90.4} & \multicolumn{1}{c}{86.0} & \multicolumn{1}{c}{88.1} & \multicolumn{1}{c}{82.7} & \multicolumn{1}{c}{78.5} & \multicolumn{1}{c}{73.0} \\ 
Boundary TextSpotter~\cite{lu2022boundary} & S-1080 & \multicolumn{1}{c}{88.7} & \multicolumn{1}{c}{84.6} & \multicolumn{1}{c}{86.6} & \multicolumn{1}{c}{82.5} & \multicolumn{1}{c}{77.4} & \multicolumn{1}{c}{71.7} \\
ABINet++~\cite{fang2022abinet++} & S-1000 & \multicolumn{1}{c}{-} & \multicolumn{1}{c}{-} & \multicolumn{1}{c}{-} & \multicolumn{1}{c}{\underline{84.1}} & \multicolumn{1}{c}{\underline{80.4}} & \multicolumn{1}{c}{\underline{75.4}} \\ 
 
\midrule
 CharNet R-50~\cite{xing2019convolutional} &- &\multicolumn{1}{c}{91.2} &\multicolumn{1}{c}{\underline{88.3}} &\multicolumn{1}{c}{\underline{89.7}} &80.1 & \multicolumn{1}{c}{74.5} &\multicolumn{1}{c}{62.2} \\
PGNet~\cite{wang2021pgnet} & L-1536 & \multicolumn{1}{c}{91.8} & \multicolumn{1}{c}{84.8} & \multicolumn{1}{c}{88.2} & \multicolumn{1}{c}{83.3} & \multicolumn{1}{c}{78.3} & \multicolumn{1}{c}{63.5} \\
MANGO*~\cite{qiao2021mango}  & L-1800 & \multicolumn{1}{c}{-} & \multicolumn{1}{c}{-}  & \multicolumn{1}{c}{-}  & \multicolumn{1}{c}{81.8} & \multicolumn{1}{c}{78.9} & \multicolumn{1}{c}{67.3} \\
\midrule     
TextFormer & S-1000 & \multicolumn{1}{c}{\textbf{94.3}} & \multicolumn{1}{c}{\textbf{89.2}} & \multicolumn{1}{c}{\textbf{91.7}} & \multicolumn{1}{c}{\textbf{84.5}} & \multicolumn{1}{c}{\textbf{80.9}} & \multicolumn{1}{c}{\textbf{76.0}} \\
\bottomrule
\end{tabular}
\end{table*}

\begin{figure*}[h!t]
\centering
\includegraphics[width=1.0\linewidth]{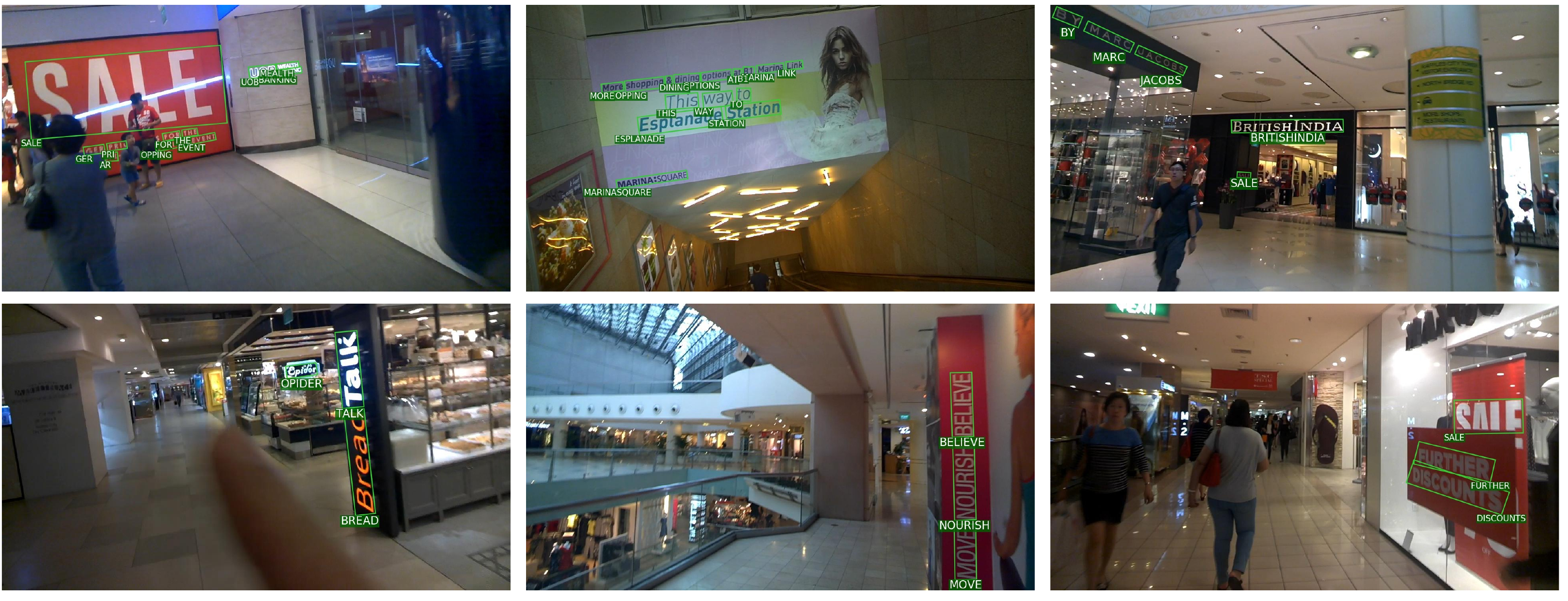}
\caption{Visualization examples for ICDAR 2015 dataset without lexicons. Best zoom in for better visualization.}
\label{fig5}
\end{figure*}

\subsubsection{Loss Function}
For the segmentation task, we use a combination of a dice loss~\cite{milletari2016v} and a focal loss~\cite{lin2017focal}. For the classification and recognition tasks, we simply use the cross-entropy loss for each predicted text instance.
Then, the overall training loss is as follows,
\begin{equation}
    \mathcal{L}= \mathcal{L}_{cls} + \lambda_{mask}(\mathcal{L}_{focal} +\mathcal{L}_{dice})+\lambda_{rec}\mathcal{L}_{rec}
\end{equation}
where $\lambda_{mask}$ and $\lambda_{rec}$ are the weights of segmentation and transcription losses.

Training our model with text annotations or weak labels, the matching cost is based solely on the classification and recognition cost. And the loss terms are composed of the classification loss and the recognition loss. There is a slight difference when training with weak labels that we only consider the classification loss between the matched query and the ground truth (1 to 1). We do this to avoid text queries from predicting only single text in a text image.

\begin{table*}[h!t]
\center
\caption{Text spotting results on Total-Text. ``None" represents lexicon-free and ``Full" means lexicons of all images. MANGO*~\cite{qiao2021mango} is evaluated with IOU 0.1. \label{tab2}}
\begin{tabular}{lllllll}
\toprule
\multirow{2}{*}{Method}	& \multirow{2}{*}{Input Size} & \multicolumn{3}{c}{Detection} & \multicolumn{2}{c}{End-to-end}  \\
\cmidrule(lr){3-5} \cmidrule(lr){6-7}
& & \multicolumn{1}{c}{Precision} & \multicolumn{1}{c}{Recall} & \multicolumn{1}{c}{F-measure} & \multicolumn{1}{c}{None} & \multicolumn{1}{c}{Full} \\
\midrule
Mask TextSpotter~\cite{lyu2018mask} & S-1000 & \multicolumn{1}{c}{69.0} & \multicolumn{1}{c}{55.0} & \multicolumn{1}{c}{61.3} & \multicolumn{1}{c}{52.9} & \multicolumn{1}{c}{71.8} \\
TextNet~\cite{sun2019textnet} & - & \multicolumn{1}{c}{68.2} & \multicolumn{1}{c}{59.5} & \multicolumn{1}{c}{63.5} & \multicolumn{1}{c}{54.0} & \multicolumn{1}{c}{-} \\
TextDragon~\cite{feng2019textdragon} & - & \multicolumn{1}{c}{85.6} & \multicolumn{1}{c}{75.5} & \multicolumn{1}{c}{80.3} & \multicolumn{1}{c}{48.8} & \multicolumn{1}{c}{74.8} \\
ABCNet ~\cite{liu2020abcnet} & - & \multicolumn{1}{c}{-} & \multicolumn{1}{c}{-} & \multicolumn{1}{c}{-} & \multicolumn{1}{c}{64.2} & \multicolumn{1}{c}{75.7} \\
ABCNet-MS ~\cite{liu2020abcnet} & - & \multicolumn{1}{c}{-} & \multicolumn{1}{c}{-} & \multicolumn{1}{c}{-} & \multicolumn{1}{c}{69.5} & \multicolumn{1}{c}{78.4} \\
Unconstrained ~\cite{qin2019towards} & S-600 & \multicolumn{1}{c}{83.3} & \multicolumn{1}{c}{83.4} & \multicolumn{1}{c}{83.3} & \multicolumn{1}{c}{67.8} & \multicolumn{1}{c}{-} \\
Text Perceptron~\cite{qiao2020text} & L-2000 & \multicolumn{1}{c}{87.5} & \multicolumn{1}{c}{81.9} & \multicolumn{1}{c}{84.6} & \multicolumn{1}{c}{57.0} & \multicolumn{1}{c}{76.6} \\
Mask TextSpotter v3~\cite{liao2020mask} & - & \multicolumn{1}{c}{-} & \multicolumn{1}{c}{-} & \multicolumn{1}{c}{-} & \multicolumn{1}{c}{71.2} & \multicolumn{1}{c}{78.4} \\
PAN++~\cite{wang2021pan++} & S-736 & \multicolumn{1}{c}{-} & \multicolumn{1}{c}{-} & \multicolumn{1}{c}{-} & \multicolumn{1}{c}{68.6} & \multicolumn{1}{c}{78.6} \\
ABCNet V2~\cite{liu2021abcnetv2} & S-1000 & \multicolumn{1}{c}{\textbf{90.2}} & \multicolumn{1}{c}{84.1} & \multicolumn{1}{c}{\underline{87.0}} & \multicolumn{1}{c}{70.4} & \multicolumn{1}{c}{78.1} \\
Boundary TextSpotter~\cite{lu2022boundary} & S-800 & \multicolumn{1}{c}{89.6} & \multicolumn{1}{c}{81.2} & \multicolumn{1}{c}{85.2} & \multicolumn{1}{c}{66.2} & \multicolumn{1}{c}{78.4} \\
ABINet++~\cite{fang2022abinet++} & S-1000 & \multicolumn{1}{c}{-} & \multicolumn{1}{c}{-} & \multicolumn{1}{c}{-} & \multicolumn{1}{c}{\underline{77.6}} & \multicolumn{1}{c}{\underline{84.5}} \\
\midrule
CharNet~\cite{xing2019convolutional} &- &\multicolumn{1}{c}{\underline{89.9}} &\multicolumn{1}{c}{81.7} &\multicolumn{1}{c}{85.6} &\multicolumn{1}{c}{66.0} &\multicolumn{1}{c}{-} \\
PGNet~\cite{wang2021pgnet} &L-640 &\multicolumn{1}{c}{86.8} &\multicolumn{1}{c}{\textbf{85.5}} &\multicolumn{1}{c}{86.1} & \multicolumn{1}{c}{63.1} &\multicolumn{1}{c}{-} \\
MAGNO*~\cite{qiao2021mango} & L-1280 & \multicolumn{1}{c}{-} & \multicolumn{1}{c}{-} & \multicolumn{1}{c}{-} & \multicolumn{1}{c}{71.7} & \multicolumn{1}{c}{82.6} \\
\midrule
TextFormer & S-1000 & \multicolumn{1}{c}{89.3} & \multicolumn{1}{c}{\underline{85.0}} & \multicolumn{1}{c}{\textbf{87.1}} & \multicolumn{1}{c}{\textbf{77.9}} &\multicolumn{1}{c}{\textbf{84.9}} \\
\bottomrule
\end{tabular}
\end{table*}

\begin{figure*}[h!t]
\centering
\includegraphics[width=0.95\linewidth]{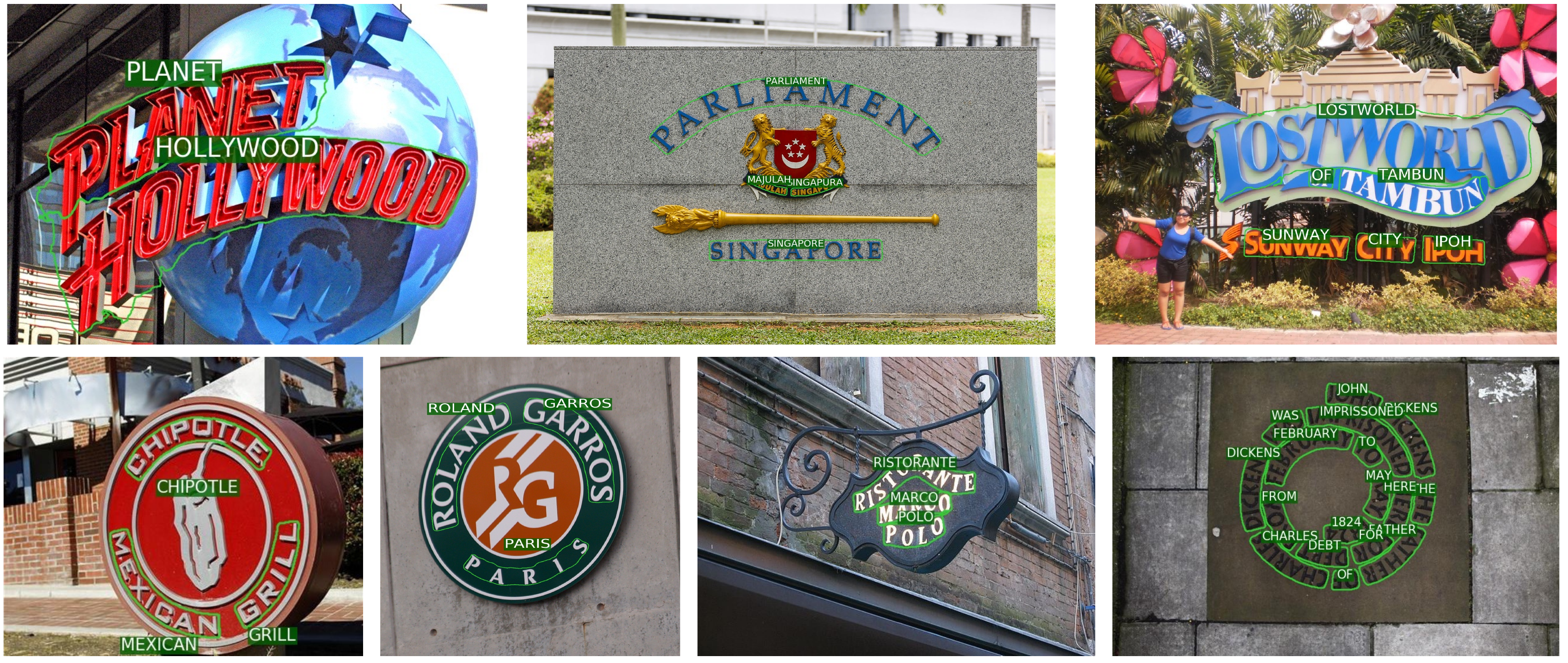}
\caption{Visualization examples for Total-Text dataset without lexicons. Zoom in for better view.}
\label{fig6}
\end{figure*}

\section{Experiments}

To evaluate the effectiveness and robustness of our proposed TextFormer, we conduct experiments on several challenging scene text benchmarks, including multi-oriented text dataset ICDAR 2015~\cite{karatzas2015icdar}, arbitrarily-shaped text dataset Total-Text~\cite{ch2017total}, two bilingual text benchmarks ReCTS~\cite{zhang2019icdar} and LSVT~\cite{sun2019icdar}, and ambiguous text dataset TDA-ReCTS~\cite{wang2020ae}.
Ablation studies are preformed on ICDAR2015 and TDA-ReCTS to
verify the legitimacy of our recognition branch and mixed supervision.

\begin{table*}[h!t]
\center
\caption{Text spotting results on TDA-ReCTS and ReCTS benchmarks. ``P", ``R", ``F", and ``1-NED" indicate precision, recall, F-measure, and normalized edit distance, respectively. Method marked with * means the model is trained with an extra weakly annotated dataset LSVT (only a single text-of-interest transcription is provided per image without any location annotation). \label{tab3}} 
\resizebox{\textwidth}{!}{
\begin{tabular}{lllllllll}
\toprule
\multirow{2}{*}{Method}	& \multicolumn{4}{c}{TDA-ReCTS} & \multicolumn{4}{c}{ReCTS}  \\
\cmidrule(lr){2-5} \cmidrule(lr){6-9} 
 & Precision & Recall & F-measure & 1-NED & Precision & Recall & F-measure & 1-NED \\
 \midrule
EAST ~\cite{zhou2017east}  & \multicolumn{1}{c}{70.6} & \multicolumn{1}{c}{61.6} & \multicolumn{1}{c}{65.8} & \multicolumn{1}{c}{-} & \multicolumn{1}{c}{74.3} & \multicolumn{1}{c}{73.7} & \multicolumn{1}{c}{74.0} & \multicolumn{1}{c}{-} \\
PSENet ~\cite{wang2019shape} & \multicolumn{1}{c}{72.1} & \multicolumn{1}{c}{65.5} & \multicolumn{1}{c}{68.7} & \multicolumn{1}{c}{-} & \multicolumn{1}{c}{87.3} & \multicolumn{1}{c}{83.9} & \multicolumn{1}{c}{85.6} & \multicolumn{1}{c}{-} \\
FOTS ~\cite{liu2018fots} & \multicolumn{1}{c}{68.2} & \multicolumn{1}{c}{70.0} & \multicolumn{1}{c}{69.1} & \multicolumn{1}{c}{34.9} & \multicolumn{1}{c}{-} & \multicolumn{1}{c}{-} & \multicolumn{1}{c}{-} & \multicolumn{1}{c}{50.8} \\
Mask TextSpotter ~\cite{lyu2018mask} & \multicolumn{1}{c}{80.1} & \multicolumn{1}{c}{74.9} & \multicolumn{1}{c}{77.4} & \multicolumn{1}{c}{46.7} & \multicolumn{1}{c}{89.3} & \multicolumn{1}{c}{88.8} & \multicolumn{1}{c}{89.0} & \multicolumn{1}{c}{67.8} \\
ABCNet v2 ~\cite{liu2021abcnetv2} & \multicolumn{1}{c}{-} & \multicolumn{1}{c}{-} & \multicolumn{1}{c}{-} & \multicolumn{1}{c}{-} & \multicolumn{1}{c}{93.6} & \multicolumn{1}{c}{87.5} & \multicolumn{1}{c}{90.4} & \multicolumn{1}{c}{62.7} \\
AE TextSpotter ~\cite{wang2020ae} & \multicolumn{1}{c}{\textbf{84.8}} & \multicolumn{1}{c}{78.3} & \multicolumn{1}{c}{81.4} & \multicolumn{1}{c}{51.3} & \multicolumn{1}{c}{92.3} & \multicolumn{1}{c}{\textbf{91.5}} & \multicolumn{1}{c}{91.9} & \multicolumn{1}{c}{73.1} \\
\midrule
TextFormer & \multicolumn{1}{c}{84.6} & \multicolumn{1}{c}{\underline{82.0}} & \multicolumn{1}{c}{\underline{83.3}} & \multicolumn{1}{c}{\underline{63.1}} & \multicolumn{1}{c}{\underline{94.2}} & \multicolumn{1}{c}{89.8} & \multicolumn{1}{c}{\underline{91.9}} & \multicolumn{1}{c}{\underline{77.7}} \\
TextFormer* & \multicolumn{1}{c}{\underline{84.6}} & \multicolumn{1}{c}{\textbf{82.7}} & \multicolumn{1}{c}{\textbf{83.6}} & \multicolumn{1}{c}{\textbf{64.5}} & \multicolumn{1}{c}{\textbf{94.4}} & \multicolumn{1}{c}{\underline{90.0}} & \multicolumn{1}{c}{\textbf{92.2}} & \multicolumn{1}{c}{\textbf{78.8}} \\
 \bottomrule
\end{tabular}}
\end{table*}

\subsection{Datasets}

\vspace{0.1cm}
\noindent
\textbf{SynthText 150k}~\cite{liu2020abcnet} is a synthesized dataset for arbitrarily-shaped scene text. It contains 150k synthetic text images with polygon annotations, which are composed of one-third of curved text, and the rest are multi-oriented text.

\vspace{0.1cm}
\noindent
\textbf{SynChinese 130k}~\cite{liu2021abcnetv2} is a synthesized dataset for bilingual scene text (English and Chinese) with a multi-oriented arrangement. It includes 130k synthetic images with quadrilateral annotations.

\vspace{0.1cm}
\noindent
\textbf{ICDAR 2015}~\cite{karatzas2015icdar} consists of 1,000 training images and 500 testing images that are incidentally captured in the scene street. It contains primarily multi-oriented and small text, which are hard to detect and recognize.

\vspace{0.1cm}
\noindent
\textbf{Total-Text}~\cite{ch2017total} is an arbitrary-shaped dataset with horizontal, multi-oriented, and curved text. It has 1,255 images for training and 300 images for testing, annotated with the polygonal bounding box at the word level. Each image has at least one curved text.

\vspace{0.1cm}
\noindent
\textbf{ReCTS}~\cite{zhang2019icdar} is a Chinese multi-orientation natural scene text dataset. It contains 25k signboard images divided into 20k images for training and 5k images for testing. Compared to the English dataset, Chinese characters have a large number of categories, and their layout and arrangement are more complex and varied.

\vspace{0.1cm}
\noindent
\textbf{TDA-ReCTS}~\cite{wang2020ae} is a multi-language validation benchmark for text detection ambiguity, which includes 1k images selected from ReCTS in the case of large character spacing or juxtaposed text lines. And the rested ReCTS images are deemed as the training set.

\vspace{0.1cm}
\noindent
\textbf{LSVT}~\cite{sun2019icdar} provides a large variety of texts from street view in which 50k images are fully annotated, and 400k images are partially annotated (only provides one transcription annotation per image).

\subsection{Implementation Details}

\subsubsection{Network}
The backbone network is ResNet-50 pre-trained on ImageNet~\cite{krizhevsky2017imagenet}, and the text query number is set to 50. The recognition branch is composed of two layers of Transformer decoder with 32 character queries for the English datasets and 50 character queries for the Chinese datasets. All models are trained with a batch size of 2 on 2 A100 GPUs, using AdamW~\cite{loshchilov2017decoupled} optimizer with poly learning rate strategy~\cite{chen2017deeplab}. The initial learning rate is $10^{-4}$ with the weight decay of $0.05$. 

\subsubsection{Training}
We divide the training process into two steps: pre-training and fine-tuning. For Chinese datasets like TDA-ReCTS and ReCTS, we pre-train our model on SynChinese 130k and fine-tune it on the target datasets. And we further train our model with mixed supervision, using the mixture of the weakly-annotated dataset LSVT and the target dataset. For English datasets like ICDAR 2015 and Total-Text, the pre-trained data is collected from public datasets, including SynthText 150k, ICDAR 2013~\cite{karatzas2013icdar}, ICDAR-MLT~\cite{nayef2017icdar2017}, COCO-Text~\cite{veit2016cocotext} and ArT~\cite{chng2019icdar2019}. And the pre-trained model is then fine-tuned on the target datasets. The following data augmentations are used in our training, which include: (1) randomly resizing the shorter side of the image of [512, 640, 800, 1024], and (2) randomly rotating the image with an angle of $[0^\circ, 90^\circ, 180^\circ, 270^\circ]$. And the size of the character dictionary is set to be 36 (including 26 letters, 10 digits) for English datasets and 5462 for Chinese datasets.

\begin{figure*}[h!t]
\centering
\includegraphics[width=1.0\linewidth]{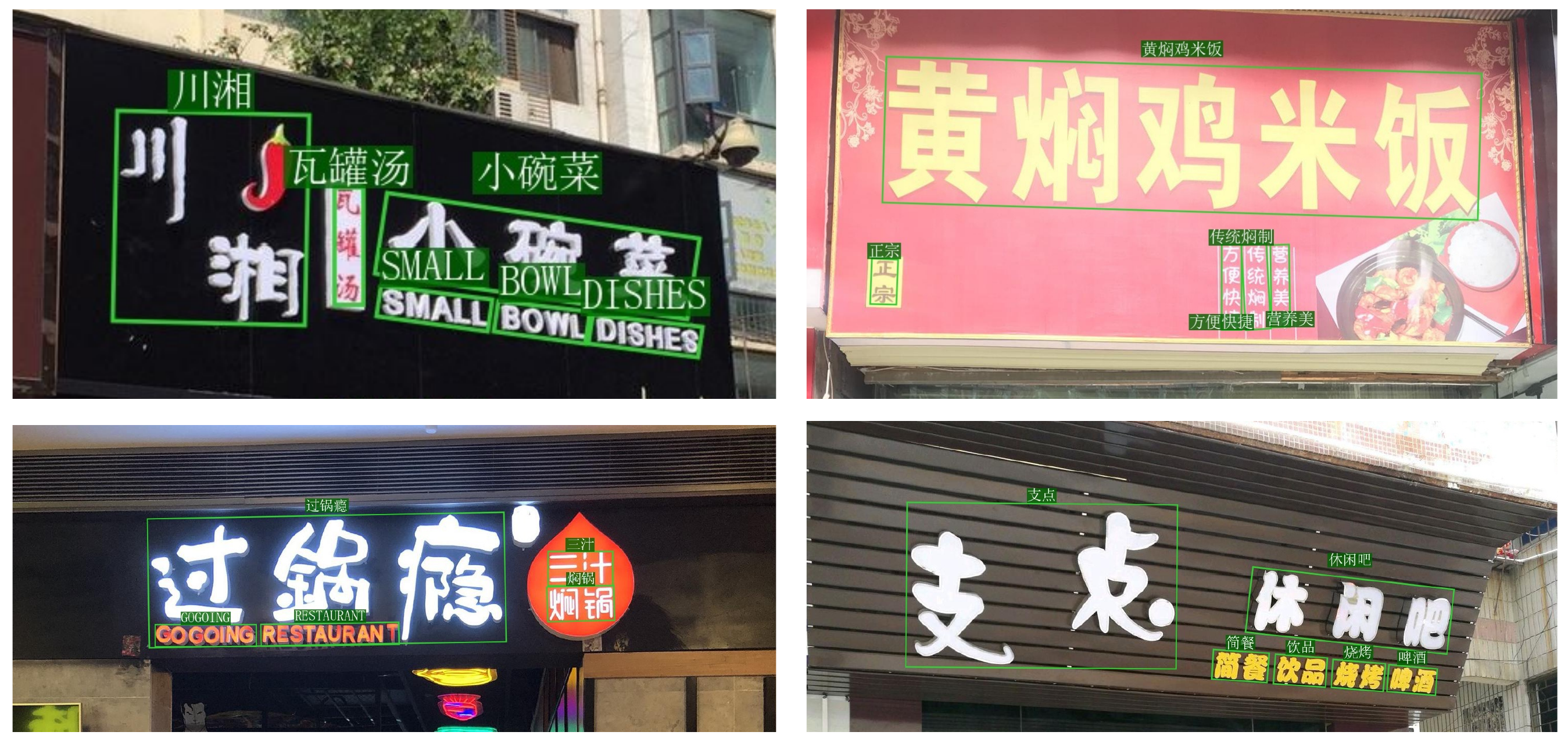}
\caption{Visualization examples for TDA-ReCTS dataset. Best zoom in for better visualization.}
\label{fig7}
\end{figure*}

\subsubsection{Inference}
During inference stage, the shorter side of the image is resized to 1000 while keeping the aspect ratio.

\subsection{Evaluation of English Text Spotting Benchmarks}
The evaluation metric follows the standard polygon evaluation principle with a 0.5 IoU threshold for a fair comparison. And the data in the table are presented in percentages.

We first test our method on ICDAR 2015 to show the effectiveness on multi-oriented scene text dataset.
Table~\ref{tab1} shows that TextFormer achieves the best performance on detection and end-to-end (E2E) items under ``Strong", ``Weak", and ``General" lexicons comparing to previous methods. The best recall result from Table~\ref{tab1} reveals the powerful ability of our model to capture the small text instance in the scene image.
And our method outperforms ABINet++ in terms of the ``General" metric without using an extra language model. Some example results for ICDAR 2015 are shown in Fig.~\ref{fig5}. Whether it is the horizontal, vertical, dense, blurred, or obscured layout, our method can successfully detect and recognize these texts successfully.

For the arbitrarily-shaped scene text, we conducted an experiment on Total-Text to test the performance of our model.
The results are presented in Table~\ref{tab2}.
In the text detection task, our model can achieve a comparable result comparing to ABCNet v2.
As for the end-to-end text spotting task, TextFormer surpasses the state-of-the-art method both in ``None" and ``Full" metrics.
Some qualitative results are shown in Fig.~\ref{fig6}.
In summary, the results on ICDAR 2015 and Total-Text demonstrate the superiority of our model for multi-oriented and arbitrarily-shaped text spotting. 

\subsection{Evaluation of Chinese Text Spotting Benchmarks}
In addition to the English text spotting dataset, we also evaluate our method on Chinese text spotting benchmarks to further verify the generalization ability of TextFormer. Compared to reading Latin-based text, Chinese text spotting remains challenging due to its large vocabulary of Chinese characters and detection ambiguity, which often occurs when multiple horizontal and vertical text lines are interspersed. Therefore, we evaluate our method on TDA-ReCTS, ReCTS, and LSVT following the standard evaluation protocol metrics~\cite{wang2020ae}.

As shown in Table~\ref{tab3}, our method achieves the best performance on TDA-ReCTS and ReCTS datasets compared to previous end-to-end methods. Especially on the TDA-ReCTS dataset, which is proposed for ambiguous text spotting, TextFormer outperforms AE TextSpotter by 11.8\% (63.1\% vs. 51.3\%) in terms of 1-NED without using an additional language model. We attribute this to the query-based multi-task modeling that makes our model focus on semantic text area rather than meaningless detection region. Furthermore, by training our model with weak annotation that contains a single text transcription per image, TextFormer* can further improve the detection and end-to-end performance which achieves an E2E 1-NED of 64.5\% on TDA-ReCTS and 78.8\% on ReCTS. This greatly validates the effectiveness of our proposed mixed-supervision learning scheme.
Some examples for TDA-ReCTS are shown in Fig.~\ref{fig7}.
Our model can handle ambiguous texts well and learn the reading order correctly. 

And we also conduct an experiment on the ICDAR 2019 LSVT competition.
Our TextFormer achieves the F-measure of 54.2\% and the 1-NED of 60.7\%.  And with the use of weak annotations provided in LSVT dataset, it can further improve end-to-end performance and achieve an E2E 1-NED of 62.2\%. Note that without using extra datasets, our single-scale and single-model result is comparable to the Top3 methods on the leader board, which using extra datasets, multi-model ensemble, and multi-scale testing strategy.

\begin{table*}[h!t]
\center
\caption{Ablation study of different tasks and different supervisions on TDA-ReCTS. The columns ``Masks" and ``Texts" mean training the model with the annotations of instance masks and texts, respectively. ``Weak Texts" denotes only one text transcription per image is used to train the model. \label{tab4}}
\resizebox{\textwidth}{!}{
\begin{tabular}{llllllllll}
\toprule
\multirow{2}{*}{Row} &\multicolumn{2}{c}{Tasks} 
&\multicolumn{3}{c}{Annotations} &\multicolumn{3}{c}{Detection} &\multicolumn{1}{c}{End-to-end}\\

\cmidrule(lr){2-3} \cmidrule(lr){4-6} \cmidrule(lr){7-9}  \cmidrule(lr){10-10}

& \multicolumn{1}{c}{Detection} & \multicolumn{1}{c}{Recognition} & \multicolumn{1}{c}{Masks} & \multicolumn{1}{c}{Texts} & \multicolumn{1}{c}{Weak Texts} & \multicolumn{1}{c}{Precision} & \multicolumn{1}{c}{Recall} &\multicolumn{1}{c}{F-measure} & \multicolumn{1}{c}{1-NED} \\
\midrule
\multicolumn{1}{c}{1} & \multicolumn{1}{c}{\checkmark} & \multicolumn{1}{c}{$\times$} & \multicolumn{1}{c}{\checkmark} & \multicolumn{1}{c}{$\times$} & \multicolumn{1}{c}{$\times$} & \multicolumn{1}{c}{83.3} & \multicolumn{1}{c}{78.2} & \multicolumn{1}{c}{80.7} & \multicolumn{1}{c}{-} \\
\multicolumn{1}{c}{2} & \multicolumn{1}{c}{$\times$} & \multicolumn{1}{c}{\checkmark} & \multicolumn{1}{c}{$\times$} & \multicolumn{1}{c}{\checkmark} & \multicolumn{1}{c}{$\times$} & \multicolumn{1}{c}{78.9} & \multicolumn{1}{c}{61.8} & \multicolumn{1}{c}{69.3} & \multicolumn{1}{c}{44.3} \\
\multicolumn{1}{c}{3} & \multicolumn{1}{c}{$\times$} & \multicolumn{1}{c}{\checkmark} & \multicolumn{1}{c}{$\times$} & \multicolumn{1}{c}{\checkmark} & \multicolumn{1}{c}{\checkmark} & \multicolumn{1}{c}{73.4} & \multicolumn{1}{c}{66.6} & \multicolumn{1}{c}{69.8} & \multicolumn{1}{c}{47.0} \\
\multicolumn{1}{c}{4} & \multicolumn{1}{c}{\checkmark} & \multicolumn{1}{c}{\checkmark} & \multicolumn{1}{c}{\checkmark} & \multicolumn{1}{c}{\checkmark} & \multicolumn{1}{c}{$\times$} & \multicolumn{1}{c}{\textbf{84.6}} & \multicolumn{1}{c}{82.0} & \multicolumn{1}{c}{83.3} & \multicolumn{1}{c}{63.1} \\
\multicolumn{1}{c}{5} & \multicolumn{1}{c}{\checkmark} & \multicolumn{1}{c}{\checkmark} & \multicolumn{1}{c}{\checkmark} & \multicolumn{1}{c}{\checkmark} & \multicolumn{1}{c}{\checkmark} & \multicolumn{1}{c}{\textbf{84.6}} & \multicolumn{1}{c}{\textbf{82.7}} & \multicolumn{1}{c}{\textbf{83.6}} & \multicolumn{1}{c}{\textbf{64.5}} \\
\bottomrule
\end{tabular}}
\end{table*}

\begin{table*}[h!t]
\center
\caption{Comparison of different feature extractors on ICDAR 2015. ``S", ``W", ``G" mean text recognition with Strong (S), Weak (W), and Generic (G) lexicon, respectively.\label{tab5}}

\begin{tabular}{llllllll}
\toprule
\multirow{2}{*}{Row} & \multirow{2}{*}{Feature Extractor} &
\multicolumn{3}{c}{Detection} & \multicolumn{3}{c}{End-to-end} \\
\cmidrule(lr){3-5} \cmidrule(lr){6-8} 
& & Precision & Recall & F-measure & S & W & G \\
\midrule
\multicolumn{1}{c}{1} & Masked RoI & \multicolumn{1}{c}{94.3} & \multicolumn{1}{c}{89.0} & \multicolumn{1}{c}{91.6} & \multicolumn{1}{c}{77.1} & \multicolumn{1}{c}{77.0} & \multicolumn{1}{c}{70.0} \\
\multicolumn{1}{c}{2} & $AGG_v$ & \multicolumn{1}{c}{93.9} & \multicolumn{1}{c}{88.4} & \multicolumn{1}{c}{91.0} & \multicolumn{1}{c}{78.5} & \multicolumn{1}{c}{78.9} & \multicolumn{1}{c}{73.8} \\
\multicolumn{1}{c}{3} & $AGG_h$ & \multicolumn{1}{c}{\textbf{95.4}} & \multicolumn{1}{c}{85.2} & \multicolumn{1}{c}{90.0} & \multicolumn{1}{c}{83.0} & \multicolumn{1}{c}{79.5} & \multicolumn{1}{c}{75.7} \\
\multicolumn{1}{c}{4} & $AGG$ & \multicolumn{1}{c}{94.3} & \multicolumn{1}{c}{\textbf{89.2}} & \multicolumn{1}{c}{\textbf{91.7}} & \multicolumn{1}{c}{\textbf{84.5}} & \multicolumn{1}{c}{\textbf{80.9}} & \multicolumn{1}{c}{\textbf{76.0}} \\
\bottomrule
\end{tabular}
\end{table*}

\subsection{Ablation Studies}

In this section, we first conduct an ablation study on TDA-ReCTS to verify the effectiveness of multi-task modeling and our proposed mixed supervision. And then, we do ablation studies on ICDAR 2015 to evaluate the effectiveness of our feature extractor and the design of the text recognizer.

\subsubsection{Effectiveness of Multi-task Modeling}
To prove the superiority of multi-task modeling, we compare the performance of a single detector (row \#1) with our end-to-end text spotter (row \#4) on TDA-ReCTS. In the spirit of equality, the detector is separated from our end-to-end text spotter. The detection results are shown in Table \ref{tab4}. The end-to-end model achieves 83.3\% F-measure on the text detection task, which surpasses the single detector in 2.6\% (83.3\% vs. 80.7\%).
It can be seen that the multi-task modeling improves the detection result greatly. The potential reason is that in the end-to-end framework, the detection branch and recognition branch can be mutually trained. They share the same backbone and semantic features, so the recognition loss can propagate to the detection branch, which improves the detection result.

\begin{table*}[h!]
\center
\caption{Effectiveness of recognition head design under different numbers of decoder layers and different recognition loss on ICDAR 2015. ``S", ``W", ``G" mean text recognition with Strong (S), Weak (W), and Generic (G) lexicon, respectively. ``P", ``R", ``F" indicate precision, recall and f-measure. ``CE" represents cross-entropy loss. \label{tab6}}

\begin{tabular}{llllllll}
\toprule
\multirow{2}{*}{Layer} & \multirow{2}{*}{Loss} &
\multicolumn{3}{c}{Detection} &
\multicolumn{3}{c}{End-to-end} \\
\cmidrule(lr){3-5} \cmidrule(lr){6-8} 
& & \multicolumn{1}{c}{P} & \multicolumn{1}{c}{R} & \multicolumn{1}{c}{F} & \multicolumn{1}{c}{S} & \multicolumn{1}{c}{W} & \multicolumn{1}{c}{G} \\       
\midrule
\multicolumn{1}{c}{1L} & CE & \multicolumn{1}{c}{\textbf{95.3}} & \multicolumn{1}{c}{86.3} & \multicolumn{1}{c}{90.6} & \multicolumn{1}{c}{82.2} & \multicolumn{1}{c}{78.6} & \multicolumn{1}{c}{72.3} \\
\multicolumn{1}{c}{2L} & CE & \multicolumn{1}{c}{94.3} & \multicolumn{1}{c}{\textbf{89.2}} & \multicolumn{1}{c}{\textbf{91.7}} & \multicolumn{1}{c}{\textbf{84.5}} & \multicolumn{1}{c}{\textbf{80.9}} & \multicolumn{1}{c}{\textbf{76.0}}\\
\multicolumn{1}{c}{2L} & CTC & \multicolumn{1}{c}{95.1} & \multicolumn{1}{c}{85.8} & \multicolumn{1}{c}{90.0} & \multicolumn{1}{c}{82.8} & \multicolumn{1}{c}{79.4} & \multicolumn{1}{c}{74.2} \\
\multicolumn{1}{c}{3L} & CE & \multicolumn{1}{c}{94.6} & \multicolumn{1}{c}{86.0} & \multicolumn{1}{c}{90.1} & \multicolumn{1}{c}{82.1} & \multicolumn{1}{c}{78.8} &  \multicolumn{1}{c}{74.3} \\
\bottomrule
\end{tabular}
\end{table*}

\begin{figure*}[h!t]
\centering
\includegraphics[width=1.0\linewidth]{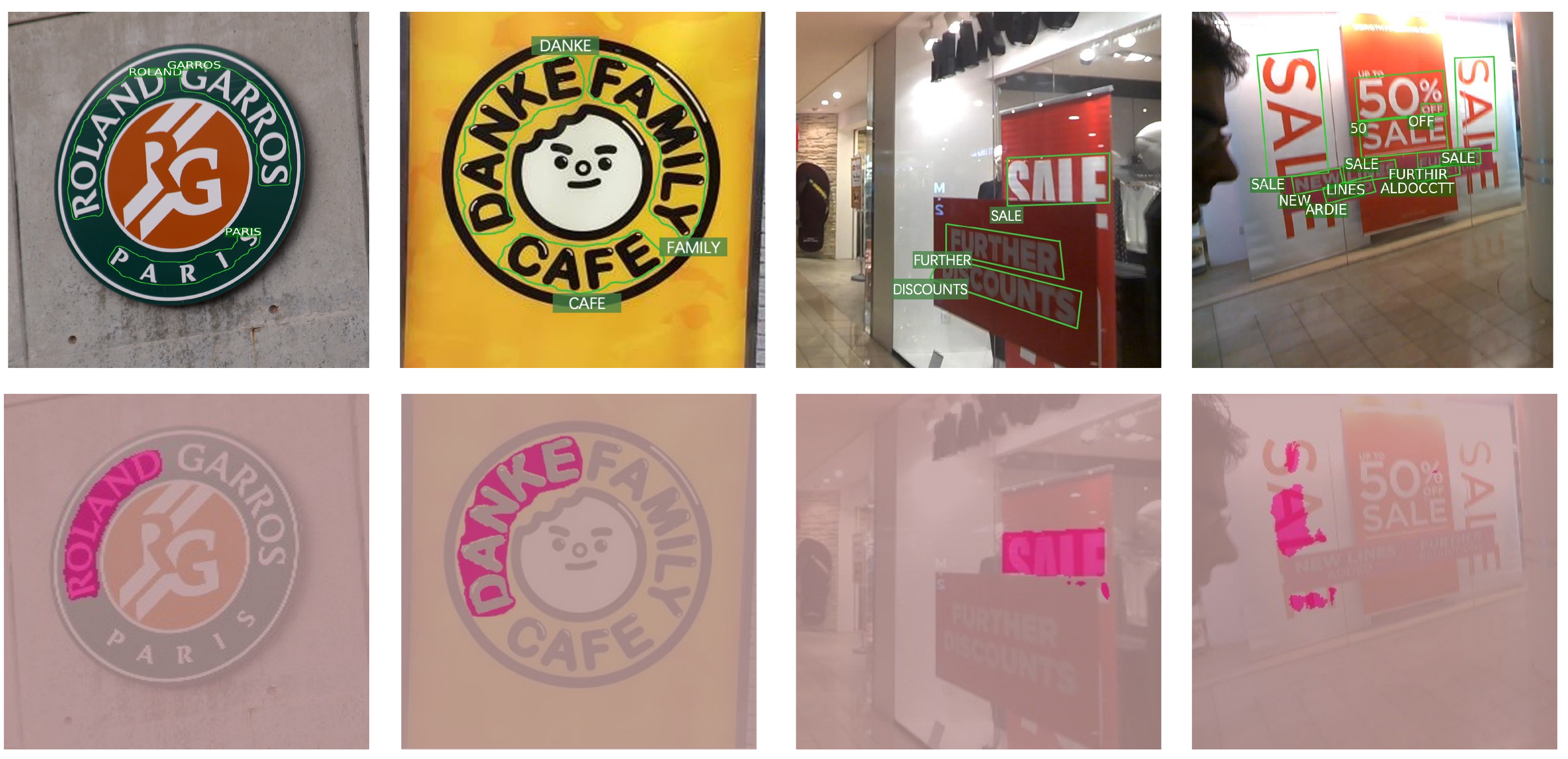}
\caption{Visualization of segmentation results (first row pictures) and attention masks of recognition branch(second row pictures). 
For the sake of clarity, the segmentation results are represented by the polygon lines.}
\label{fig8}
\end{figure*}

\subsubsection{Effectiveness of Mixed Supervision}
As discussed in Sec \ref{mixed}, our model can be trained in a mixed-supervision learning manner. 
In Table~\ref{tab4}, we show the detection and end-to-end results on TDA-ReCTS. Using real data with only text annotations and synthetic data with full annotations (row \#2), our model can achieve a competitive result compared to previous fully supervised methods~\cite{liu2018fots, lyu2018mask}. 
And we further train our model (row \#3) using weak texts in the LSVT dataset, which only contains single transcription per image.
We see that it obtains better detection and end-to-end results.
The same improvement is shown in (row \#5).
Compared with the standard text spotter that uses full annotations to train the model (row \#4), we add additional data with weak text annotations (row \#5) which achieves a 1.4\% boost on the E2E item.
With this learning way, we can significantly reduce the labeling cost while improving the results of the model.

\subsubsection{Effectiveness of AGG Module}
As presented in Table \ref{tab6}, we show the effectiveness of our proposed AGG (row \#1 vs. row \#4). Higher detection and end-to-end results are obtained in the presence of AGG module.
Compared with Masked RoI feature extractor~\cite{wang2021pan++}, which extracts the feature locally, the model with AGG module outperforms it by 6\% in ``General" metric.
This demonstrates that our recognition head benefits a lot from the proposed global feature extractor.
And we also do ablation studies on $AGG_v$ and $AGG_h$, which contain single-orientation information.
The results are shown on (Row \#2) and (Row\#3) in Table~\ref{tab6}, we observe that competitive results can still be achieved.

\subsubsection{Influence of the Number of Decoder Layers and Recognition Loss}
To verify the legitimacy of our recognition head, we do ablation studies on recognition head and recognition loss.
The results are shown in Table \ref{tab5}.
We note that using just one layer of decoder, the model can still achieve 72.3 \% on ``General" item. 
As the number of layers increases, the amount of data and training time for the convergence of recognition head are enhanced.
And the model with CE loss performs better than CTC~\cite{graves2012connectionist} loss with 1.8\% improvement.
For data and performance considerations, we choose two layers of decoder and CE loss as our recognition head.

\subsection{Visualization Analysis}
With the help of multi-task modeling, our method generates high-quality polygons and transcriptions for arbitrarily-shaped texts. 
It works well for vertical and curved texts, as shown in Fig.~\ref{fig6} and Fig.~\ref{fig7}.
Even for some extreme cases, like highly-occluded text, our method can still rightly recognize them by the proposed AGG module to make full use of the global context rather than local features. 
These visualization results confirm the superiority of multi-task modeling and mixed-supervision learning manner.

Fig.~\ref{fig8} shows the final segmentation results and attention masks which are generated by Eq.~\eqref{eq1}. The visual results indicate that our segmentation results and attention masks can focus on the right place. Note that in the third and fourth columns, the word ``SALE" is obscured by the other billboard. Our method can still accurately recognize the obscured word. Especially on the fourth column, the character ``E" is lost in the segmentation results while being activated in the attention mask, which means our model could predict correct transcription even with the inaccurate segmentation result. We attribute this to our multi-mask modeling design, which alleviates the dependency of the text detection and recognition parts to some extent.

\section{Conclusion and Future work}
In this paper, we propose a fully end-to-end arbitrary-shaped text spotter based on transformer. 
By modeling classification, segmentation and recognition tasks in a parallel query-based way, our model can be jointly trained and optimized. And we introduce a light-weight recognition head with AGG module to extract features globally. Without bells and whistles, our method achieves competitive results on both regular text and curved text benchmarks.

Since our model is built upon transformer architecture, its final performance is heavily depend on the scale of training data.
In the future work, we may solve this problem by self-supervised learning.

{\small
\bibliographystyle{ieee_fullname}
\bibliography{egbib}

\begin{thebibliography}{10}\itemsep=-1pt

\bibitem{antol2015vqa}
Stanislaw Antol, Aishwarya Agrawal, Jiasen Lu, Margaret Mitchell, Dhruv Batra,
  C~Lawrence Zitnick, and Devi Parikh.
\newblock Vqa: Visual question answering.
\newblock In {\em Proceedings of the IEEE international conference on computer
  vision}, pages 2425--2433, 2015.

\bibitem{bookstein1989principal}
Fred~L. Bookstein.
\newblock Principal warps: Thin-plate splines and the decomposition of
  deformations.
\newblock {\em IEEE Transactions on pattern analysis and machine intelligence},
  11(6):567--585, 1989.

\bibitem{bovzivc2021mixed}
Jakob Bo{\v{z}}i{\v{c}}, Domen Tabernik, and Danijel Sko{\v{c}}aj.
\newblock Mixed supervision for surface-defect detection: From weakly to fully
  supervised learning.
\newblock {\em Computers in Industry}, 129:103459, 2021.

\bibitem{busta2017deep}
Michal Busta, Lukas Neumann, and Jiri Matas.
\newblock Deep textspotter: An end-to-end trainable scene text localization and
  recognition framework.
\newblock In {\em Proceedings of the IEEE international conference on computer
  vision}, pages 2204--2212, 2017.

\bibitem{carion2020end}
Nicolas Carion, Francisco Massa, Gabriel Synnaeve, Nicolas Usunier, Alexander
  Kirillov, and Sergey Zagoruyko.
\newblock End-to-end object detection with transformers.
\newblock In {\em European Conference on Computer Vision}, pages 213--229.
  Springer, 2020.

\bibitem{chen2017deeplab}
Liang-Chieh Chen, George Papandreou, Iasonas Kokkinos, Kevin Murphy, and Alan~L
  Yuille.
\newblock Deeplab: Semantic image segmentation with deep convolutional nets,
  atrous convolution, and fully connected crfs.
\newblock {\em IEEE transactions on pattern analysis and machine intelligence},
  40(4):834--848, 2017.

\bibitem{cheng2022masked}
Bowen Cheng, Ishan Misra, Alexander~G Schwing, Alexander Kirillov, and Rohit
  Girdhar.
\newblock Masked-attention mask transformer for universal image segmentation.
\newblock In {\em Proceedings of the IEEE/CVF Conference on Computer Vision and
  Pattern Recognition}, pages 1290--1299, 2022.

\bibitem{ch2017total}
Chee~Kheng Ch'ng and Chee~Seng Chan.
\newblock Total-text: A comprehensive dataset for scene text detection and
  recognition.
\newblock In {\em 2017 14th IAPR International Conference on Document Analysis
  and Recognition (ICDAR)}, volume~1, pages 935--942. IEEE, 2017.

\bibitem{chng2019icdar2019}
Chee~Kheng Chng, Yuliang Liu, Yipeng Sun, Chun~Chet Ng, Canjie Luo, Zihan Ni,
  ChuanMing Fang, Shuaitao Zhang, Junyu Han, Errui Ding, et~al.
\newblock Icdar2019 robust reading challenge on arbitrary-shaped text-rrc-art.
\newblock In {\em 2019 International Conference on Document Analysis and
  Recognition (ICDAR)}, pages 1571--1576. IEEE, 2019.

\bibitem{datta2008image}
Ritendra Datta, Dhiraj Joshi, Jia Li, and James~Z Wang.
\newblock Image retrieval: Ideas, influences, and trends of the new age.
\newblock {\em ACM Computing Surveys (Csur)}, 40(2):1--60, 2008.

\bibitem{fang2022abinet++}
Shancheng Fang, Zhendong Mao, Hongtao Xie, Yuxin Wang, Chenggang Yan, and
  Yongdong Zhang.
\newblock Abinet++: Autonomous, bidirectional and iterative language modeling
  for scene text spotting.
\newblock {\em IEEE Transactions on Pattern Analysis and Machine Intelligence},
  2022.

\bibitem{feng2019textdragon}
Wei Feng, Wenhao He, Fei Yin, Xu-Yao Zhang, and Cheng-Lin Liu.
\newblock Textdragon: An end-to-end framework for arbitrary shaped text
  spotting.
\newblock In {\em Proceedings of the IEEE/CVF International Conference on
  Computer Vision}, pages 9076--9085, 2019.

\bibitem{gomez2017textproposals}
Llu{\'\i}s G{\'o}mez and Dimosthenis Karatzas.
\newblock Textproposals: a text-specific selective search algorithm for word
  spotting in the wild.
\newblock {\em Pattern recognition}, 70:60--74, 2017.

\bibitem{graves2012connectionist}
Alex Graves and Alex Graves.
\newblock Connectionist temporal classification.
\newblock {\em Supervised sequence labelling with recurrent neural networks},
  pages 61--93, 2012.

\bibitem{he2017mask}
Kaiming He, Georgia Gkioxari, Piotr Doll{\'a}r, and Ross Girshick.
\newblock Mask r-cnn.
\newblock In {\em Proceedings of the IEEE international conference on computer
  vision}, pages 2961--2969, 2017.

\bibitem{he2018end}
Tong He, Zhi Tian, Weilin Huang, Chunhua Shen, Yu Qiao, and Changming Sun.
\newblock An end-to-end textspotter with explicit alignment and attention.
\newblock In {\em Proceedings of the IEEE conference on computer vision and
  pattern recognition}, pages 5020--5029, 2018.

\bibitem{jaderberg2016reading}
Max Jaderberg, Karen Simonyan, Andrea Vedaldi, and Andrew Zisserman.
\newblock Reading text in the wild with convolutional neural networks.
\newblock {\em International journal of computer vision}, 116:1--20, 2016.

\bibitem{karatzas2015icdar}
Dimosthenis Karatzas, Lluis Gomez-Bigorda, Anguelos Nicolaou, Suman Ghosh,
  Andrew Bagdanov, Masakazu Iwamura, Jiri Matas, Lukas Neumann,
  Vijay~Ramaseshan Chandrasekhar, Shijian Lu, et~al.
\newblock Icdar 2015 competition on robust reading.
\newblock In {\em 2015 13th International Conference on Document Analysis and
  Recognition (ICDAR)}, pages 1156--1160. IEEE, 2015.

\bibitem{karatzas2013icdar}
Dimosthenis Karatzas, Faisal Shafait, Seiichi Uchida, Masakazu Iwamura,
  Lluis~Gomez i Bigorda, Sergi~Robles Mestre, Joan Mas, David~Fernandez Mota,
  Jon~Almazan Almazan, and Lluis~Pere De~Las~Heras.
\newblock Icdar 2013 robust reading competition.
\newblock In {\em 2013 12th International Conference on Document Analysis and
  Recognition}, pages 1484--1493. IEEE, 2013.

\bibitem{krizhevsky2017imagenet}
Alex Krizhevsky, Ilya Sutskever, and Geoffrey~E Hinton.
\newblock Imagenet classification with deep convolutional neural networks.
\newblock {\em Communications of the ACM}, 60(6):84--90, 2017.

\bibitem{li2017towards}
Hui Li, Peng Wang, and Chunhua Shen.
\newblock Towards end-to-end text spotting with convolutional recurrent neural
  networks.
\newblock In {\em Proceedings of the IEEE international conference on computer
  vision}, pages 5238--5246, 2017.

\bibitem{li2021structext}
Yulin Li, Yuxi Qian, Yuechen Yu, Xiameng Qin, Chengquan Zhang, Yan Liu, Kun
  Yao, Junyu Han, Jingtuo Liu, and Errui Ding.
\newblock Structext: Structured text understanding with multi-modal
  transformers.
\newblock In {\em Proceedings of the 29th ACM International Conference on
  Multimedia}, pages 1912--1920, 2021.

\bibitem{liao2020mask}
Minghui Liao, Guan Pang, Jing Huang, Tal Hassner, and Xiang Bai.
\newblock Mask textspotter v3: Segmentation proposal network for robust scene
  text spotting.
\newblock In {\em Proceedings of the European Conference on Computer Vision
  (ECCV)}, pages 706--722, 2020.

\bibitem{lin2017feature}
Tsung-Yi Lin, Piotr Doll{\'a}r, Ross Girshick, Kaiming He, Bharath Hariharan,
  and Serge Belongie.
\newblock Feature pyramid networks for object detection.
\newblock In {\em Proceedings of the IEEE conference on computer vision and
  pattern recognition}, pages 2117--2125, 2017.

\bibitem{lin2017focal}
Tsung-Yi Lin, Priya Goyal, Ross Girshick, Kaiming He, and Piotr Doll{\'a}r.
\newblock Focal loss for dense object detection.
\newblock In {\em Proceedings of the IEEE international conference on computer
  vision}, pages 2980--2988, 2017.

\bibitem{liu2018fots}
Xuebo Liu, Ding Liang, Shi Yan, Dagui Chen, Yu Qiao, and Junjie Yan.
\newblock Fots: Fast oriented text spotting with a unified network.
\newblock In {\em Proceedings of the IEEE conference on computer vision and
  pattern recognition}, pages 5676--5685, 2018.

\bibitem{liu2020abcnet}
Yuliang Liu, Hao Chen, Chunhua Shen, Tong He, Lianwen Jin, and Liangwei Wang.
\newblock Abcnet: Real-time scene text spotting with adaptive bezier-curve
  network.
\newblock In {\em Proceedings of the IEEE/CVF Conference on Computer Vision and
  Pattern Recognition}, pages 9809--9818, 2020.

\bibitem{liu2021abcnetv2}
Yuliang Liu, Chunhua Shen, Lianwen Jin, Tong He, Peng Chen, Chongyu Liu, and
  Hao Chen.
\newblock Abcnet v2: Adaptive bezier-curve network for real-time end-to-end
  text spotting.
\newblock {\em arXiv preprint arXiv:2105.03620}, 2021.

\bibitem{loshchilov2017decoupled}
Ilya Loshchilov and Frank Hutter.
\newblock Decoupled weight decay regularization.
\newblock {\em arXiv preprint arXiv:1711.05101}, 2017.

\bibitem{lu2022boundary}
Pu Lu, Hao Wang, Shenggao Zhu, Jing Wang, Xiang Bai, and Wenyu Liu.
\newblock Boundary textspotter: Toward arbitrary-shaped scene text spotting.
\newblock {\em IEEE Transactions on Image Processing}, 31:6200--6212, 2022.

\bibitem{lyu2018mask}
Pengyuan Lyu, Minghui Liao, Cong Yao, Wenhao Wu, and Xiang Bai.
\newblock Mask textspotter: An end-to-end trainable neural network for spotting
  text with arbitrary shapes.
\newblock In {\em Proceedings of the European Conference on Computer Vision
  (ECCV)}, pages 67--83, 2018.

\bibitem{milletari2016v}
Fausto Milletari, Nassir Navab, and Seyed-Ahmad Ahmadi.
\newblock V-net: Fully convolutional neural networks for volumetric medical
  image segmentation.
\newblock In {\em 2016 fourth international conference on 3D vision (3DV)},
  pages 565--571. Ieee, 2016.

\bibitem{mlynarski2019deep}
Pawel Mlynarski, Herv{\'e} Delingette, Antonio Criminisi, and Nicholas Ayache.
\newblock Deep learning with mixed supervision for brain tumor segmentation.
\newblock {\em Journal of Medical Imaging}, 6(3):034002--034002, 2019.

\bibitem{nayef2017icdar2017}
Nibal Nayef, Fei Yin, Imen Bizid, Hyunsoo Choi, Yuan Feng, Dimosthenis
  Karatzas, Zhenbo Luo, Umapada Pal, Christophe Rigaud, Joseph Chazalon, et~al.
\newblock Icdar2017 robust reading challenge on multi-lingual scene text
  detection and script identification-rrc-mlt.
\newblock In {\em 2017 14th IAPR International Conference on Document Analysis
  and Recognition (ICDAR)}, volume~1, pages 1454--1459. IEEE, 2017.

\bibitem{neumann2015real}
Luk{\'a}{\v{s}} Neumann and Ji{\v{r}}{\'\i} Matas.
\newblock Real-time lexicon-free scene text localization and recognition.
\newblock {\em IEEE transactions on pattern analysis and machine intelligence},
  38(9):1872--1885, 2015.

\bibitem{peng2022spts}
Dezhi Peng, Xinyu Wang, Yuliang Liu, Jiaxin Zhang, Mingxin Huang, Songxuan Lai,
  Jing Li, Shenggao Zhu, Dahua Lin, Chunhua Shen, et~al.
\newblock Spts: single-point text spotting.
\newblock In {\em Proceedings of the 30th ACM International Conference on
  Multimedia}, pages 4272--4281, 2022.

\bibitem{qiao2021mango}
Liang Qiao, Ying Chen, Zhanzhan Cheng, Yunlu Xu, Yi Niu, Shiliang Pu, and Fei
  Wu.
\newblock Mango: A mask attention guided one-stage scene text spotter.
\newblock In {\em Proceedings of the AAAI Conference on Artificial
  Intelligence}, volume~35, pages 2467--2476, 2021.

\bibitem{qiao2020text}
Liang Qiao, Sanli Tang, Zhanzhan Cheng, Yunlu Xu, Yi Niu, Shiliang Pu, and Fei
  Wu.
\newblock Text perceptron: Towards end-to-end arbitrary-shaped text spotting.
\newblock In {\em Proceedings of the AAAI Conference on Artificial
  Intelligence}, volume~34, pages 11899--11907, 2020.

\bibitem{qin2019towards}
Siyang Qin, Alessandro Bissacco, Michalis Raptis, Yasuhisa Fujii, and Ying
  Xiao.
\newblock Towards unconstrained end-to-end text spotting.
\newblock In {\em Proceedings of the IEEE/CVF International Conference on
  Computer Vision}, pages 4704--4714, 2019.

\bibitem{raisi2021transformer}
Zobeir Raisi, Mohamed~A Naiel, Georges Younes, Steven Wardell, and John~S
  Zelek.
\newblock Transformer-based text detection in the wild.
\newblock In {\em Proceedings of the IEEE/CVF Conference on Computer Vision and
  Pattern Recognition}, pages 3162--3171, 2021.

\bibitem{reddy2020text}
Harita Reddy, Namratha Raj, Manali Gala, and Annappa Basava.
\newblock Text-mining-based fake news detection using ensemble methods.
\newblock {\em International Journal of Automation and Computing},
  17(2):210--221, 2020.

\bibitem{ren2015faster}
Shaoqing Ren, Kaiming He, Ross Girshick, and Jian Sun.
\newblock Faster r-cnn: Towards real-time object detection with region proposal
  networks.
\newblock {\em Advances in neural information processing systems}, 28:91--99,
  2015.

\bibitem{ricoeur1971model}
Paul Ricoeur.
\newblock The model of the text: Meaningful action considered as a text.
\newblock {\em Social research}, pages 529--562, 1971.

\bibitem{rong2016guided}
Xuejian Rong, Bing Li, J~Pablo Munoz, Jizhong Xiao, Aries Arditi, and Yingli
  Tian.
\newblock Guided text spotting for assistive blind navigation in unfamiliar
  indoor environments.
\newblock In {\em Advances in Visual Computing: 12th International Symposium,
  ISVC 2016, Las Vegas, NV, USA, December 12-14, 2016, Proceedings, Part II
  12}, pages 11--22. Springer, 2016.

\bibitem{sun2019icdar}
Yipeng Sun, Zihan Ni, Chee-Kheng Chng, Yuliang Liu, Canjie Luo, Chun~Chet Ng,
  Junyu Han, Errui Ding, Jingtuo Liu, Dimosthenis Karatzas, et~al.
\newblock Icdar 2019 competition on large-scale street view text with partial
  labeling-rrc-lsvt.
\newblock In {\em 2019 International Conference on Document Analysis and
  Recognition (ICDAR)}, pages 1557--1562. IEEE, 2019.

\bibitem{sun2019textnet}
Yipeng Sun, Chengquan Zhang, Zuming Huang, Jiaming Liu, Junyu Han, and Errui
  Ding.
\newblock Textnet: Irregular text reading from images with an end-to-end
  trainable network.
\newblock In {\em Computer Vision--ACCV 2018: 14th Asian Conference on Computer
  Vision, Perth, Australia, December 2--6, 2018, Revised Selected Papers, Part
  III 14}, pages 83--99. Springer, 2019.

\bibitem{tang2022you}
Jingqun Tang, Su Qiao, Benlei Cui, Yuhang Ma, Sheng Zhang, and Dimitrios
  Kanoulas.
\newblock You can even annotate text with voice: Transcription-only-supervised
  text spotting.
\newblock In {\em Proceedings of the 30th ACM International Conference on
  Multimedia}, pages 4154--4163, 2022.

\bibitem{tang2022few}
Jingqun Tang, Wenqing Zhang, Hongye Liu, MingKun Yang, Bo Jiang, Guanglong Hu,
  and Xiang Bai.
\newblock Few could be better than all: Feature sampling and grouping for scene
  text detection.
\newblock In {\em Proceedings of the IEEE/CVF Conference on Computer Vision and
  Pattern Recognition}, pages 4563--4572, 2022.

\bibitem{tian2017wetext}
Shangxuan Tian, Shijian Lu, and Chongshou Li.
\newblock Wetext: Scene text detection under weak supervision.
\newblock In {\em Proceedings of the IEEE International Conference on Computer
  Vision}, pages 1492--1500, 2017.

\bibitem{veit2016cocotext}
Andreas Veit, Tomas Matera, Lukas Neumann, Jiri Matas, and Serge Belongie.
\newblock Coco-text: Dataset and benchmark for text detection and recognition
  in natural images.
\newblock In {\em arXiv preprint arXiv:1601.07140}, 2016.

\bibitem{wang2020all}
Hao Wang, Pu Lu, Hui Zhang, Mingkun Yang, Xiang Bai, Yongchao Xu, Mengchao He,
  Yongpan Wang, and Wenyu Liu.
\newblock All you need is boundary: Toward arbitrary-shaped text spotting.
\newblock In {\em Proceedings of the AAAI Conference on Artificial
  Intelligence}, volume~34, pages 12160--12167, 2020.

\bibitem{wang2021pgnet}
Pengfei Wang, Chengquan Zhang, Fei Qi, Shanshan Liu, Xiaoqiang Zhang, Pengyuan
  Lyu, Junyu Han, Jingtuo Liu, Errui Ding, and Guangming Shi.
\newblock Pgnet: Real-time arbitrarily-shaped text spotting with point
  gathering network.
\newblock {\em arXiv preprint arXiv:2104.05458}, 2021.

\bibitem{wang2020ae}
Wenhai Wang, Xuebo Liu, Xiaozhong Ji, Enze Xie, Ding Liang, ZhiBo Yang, Tong
  Lu, Chunhua Shen, and Ping Luo.
\newblock Ae textspotter: Learning visual and linguistic representation for
  ambiguous text spotting.
\newblock In {\em Computer Vision--ECCV 2020: 16th European Conference,
  Glasgow, UK, August 23--28, 2020, Proceedings, Part XIV 16}, pages 457--473.
  Springer, 2020.

\bibitem{wang2019shape}
Wenhai Wang, Enze Xie, Xiang Li, Wenbo Hou, Tong Lu, Gang Yu, and Shuai Shao.
\newblock Shape robust text detection with progressive scale expansion network.
\newblock In {\em Proceedings of the IEEE/CVF conference on computer vision and
  pattern recognition}, pages 9336--9345, 2019.

\bibitem{wang2021pan++}
Wenhai Wang, Enze Xie, Xiang Li, Xuebo Liu, Ding Liang, Yang Zhibo, Tong Lu,
  and Chunhua Shen.
\newblock Pan++: Towards efficient and accurate end-to-end spotting of
  arbitrarily-shaped text.
\newblock {\em IEEE Transactions on Pattern Analysis and Machine Intelligence},
  2021.

\bibitem{wu2019editing}
Liang Wu, Chengquan Zhang, Jiaming Liu, Junyu Han, Jingtuo Liu, Errui Ding, and
  Xiang Bai.
\newblock Editing text in the wild.
\newblock In {\em Proceedings of the 27th ACM international conference on
  multimedia}, pages 1500--1508, 2019.

\bibitem{xing2019convolutional}
Linjie Xing, Zhi Tian, Weilin Huang, and Matthew~R Scott.
\newblock Convolutional character networks.
\newblock In {\em Proceedings of the IEEE/CVF International Conference on
  Computer Vision}, pages 9126--9136, 2019.

\bibitem{yu2023structextv2}
Yuechen Yu, Yulin Li, Chengquan Zhang, Xiaoqiang Zhang, Zengyuan Guo, Xiameng
  Qin, Kun Yao, Junyu Han, Errui Ding, and Jingdong Wang.
\newblock Structextv2: Masked visual-textual prediction for document image
  pre-training.
\newblock {\em arXiv preprint arXiv:2303.00289}, 2023.

\bibitem{zhai2023fast}
Mingliang Zhai, Yulin Li, Xiameng Qin, Chen Yi, Qunyi Xie, Chengquan Zhang, Kun
  Yao, Yuwei Wu, and Yunde Jia.
\newblock Fast-structext: An efficient hourglass transformer with
  modality-guided dynamic token merge for document understanding.
\newblock {\em arXiv preprint arXiv:2305.11392}, 2023.

\bibitem{zhang2019icdar}
Rui Zhang, Yongsheng Zhou, Qianyi Jiang, Qi Song, Nan Li, Kai Zhou, Lei Wang,
  Dong Wang, Minghui Liao, Mingkun Yang, et~al.
\newblock Icdar 2019 robust reading challenge on reading chinese text on
  signboard.
\newblock In {\em 2019 international conference on document analysis and
  recognition (ICDAR)}, pages 1577--1581. IEEE, 2019.

\bibitem{zhou2017east}
Xinyu Zhou, Cong Yao, He Wen, Yuzhi Wang, Shuchang Zhou, Weiran He, and Jiajun
  Liang.
\newblock East: an efficient and accurate scene text detector.
\newblock In {\em Proceedings of the IEEE conference on Computer Vision and
  Pattern Recognition}, pages 5551--5560, 2017.

\bibitem{zhu2020deformable}
Xizhou Zhu, Weijie Su, Lewei Lu, Bin Li, Xiaogang Wang, and Jifeng Dai.
\newblock Deformable detr: Deformable transformers for end-to-end object
  detection.
\newblock {\em arXiv preprint arXiv:2010.04159}, 2020.

\end{thebibliography}
}

\end{document}